\documentclass{llncs}
\usepackage{graphicx}
\usepackage{amsmath,amssymb} 
\usepackage{color}
\usepackage[width=122mm,left=12mm,paperwidth=146mm,height=193mm,top=12mm,paperheight=217mm]{geometry}

\usepackage{xspace}
\usepackage{color}
\usepackage{booktabs}
\usepackage{subcaption}
\usepackage{ifthen}
\usepackage{microtype}
\usepackage{enumitem}
\setitemize[0]{leftmargin=10pt}

\usepackage{cancel}
\usepackage{soul}
\setstcolor{blue}
\usepackage{subfiles}

\usepackage{floatrow}
\newfloatcommand{capbtabbox}{table}[][\FBwidth]

\usepackage{colortbl}

\newcommand{\eg}{e.g.\@}
\newcommand{\etal}{et al.\@}

\newcommand{\ie}{i.e.\@}
\newcommand{\cf}{cf.\@}

\newlength\tikzfigwidth
\newlength\tikzfigheight

\def\realspace{\mathbb{R}} 

\newcommand{\mat}[1]{\ensuremath{\mathbf{#1}}}
 
\newcommand{\chis}[2][non]{\ensuremath{\chi^2 \ifthenelse{\equal{#1}{non}}{}{ \left(#1,#2\right)}}} 
\newcommand{\Gammaf}[1][non]{\ensuremath{\Gamma\ifthenelse{\equal{#1}{non}}{}{ \left( #1 \right)}}} 

\newcommand{\degree}[1][non]{\ensuremath{\ifthenelse{\equal{#1}{non}}{^\circ}{#1^\circ}}} 

\def\image{I}
\def\imh{h_{\image}}
\def\imw{w_{\image}}

\def\feat{F}
\def\feath{h_{\feat}}
\def\featw{w_{\feat}}
\def\featd{d_{\feat}}

\def\memory{M}

\def\flow{D}

\def\wgtsMemory{\alpha^{\memory}}
\def\wgtsFeat{\alpha^{\feat}}

\def\paramsFeat{\Theta_{\feat}}
\def\paramsFlownet{\Theta_{\flow}}
\def\paramsWgtMemory{\Psi_{\memory}}
\def\paramsWgtFeat{\Psi_{\feat}}

\def\numTimeScales{K}

\def\ncls{C}

\def\bilinearSym{\phi}

\def\aggregateSym{\psi}

\makeatletter
\renewcommand{\paragraph}{%
  \@startsection{paragraph}{4}%
  {\z@}{0.5ex \@plus 1ex \@minus .2ex}{-1em}%
  {\normalfont\normalsize\bfseries}%
}
\makeatother

\newcommand\blfootnote[1]{%
  \begingroup
  \renewcommand\thefootnote{}\footnote{#1}%
  \addtocounter{footnote}{-1}%
  \endgroup
}

\begin{document}

\pagestyle{headings}
\mainmatter
\def\ECCV18SubNumber{***}  

\title{Memory Warps for Learning Long-Term \\ Online Video Representations}

\titlerunning{Memory Warps for Learning Long-Term Online Video Representations}
\authorrunning{Tuan-Hung Vu, Wongun Choi, Samuel Schulter and Manmohan Chandraker}
\author{Tuan-Hung Vu$^{1}$ $\;$ Wongun Choi$^2$ $\;$ Samuel Schulter$^3$ $\;$ Manmohan Chandraker$^{3,4}$}
\institute{INRIA/ENS WILLOW$^1$ $\quad$ ISEE$^2$ $\quad$ NEC-Labs$^3$ $\quad$ UC San Diego$^4$}

\maketitle

\begin{abstract}

  This paper proposes a novel memory-based online video representation that is efficient, accurate and predictive. This is in contrast to prior works that often rely on computationally heavy 3D convolutions, ignore actual motion when aligning features over time, or operate in an off-line mode to utilize future frames. In particular, our memory (i) holds the feature representation, (ii) is spatially warped over time to compensate for observer and scene motions, (iii) can carry long-term information, and (iv) enables predicting feature representations in future frames. By exploring a variant that operates at multiple temporal scales, we efficiently learn across even longer time horizons. We apply our online framework to object detection in videos, obtaining a large $2.3$ times speed-up and losing only $0.9\%$ mAP on ImageNet-VID dataset, compared to prior works that even use future frames.  Finally, we demonstrate the predictive property of our representation in two novel detection setups, where features are propagated over time to (i) significantly enhance a real-time detector by more than $10\%$ mAP in a multi-threaded online setup and to (ii) anticipate objects in future frames.

\end{abstract}


\blfootnote{The work was conducted as part of Tuan-Hung Vu's internship at NEC Labs America.}
\section{Introduction}
\label{sec:intro}
Motion is a crucial intermediary for human visual perception to learn about its environment and relate to it \cite{Cutting_1986,Gibson_1979}.  By encapsulating motion cues, video represents a rich medium for computer vision to understand and analyze the visual world.  While the advent of convolutional neural networks (CNNs) has led to rapid improvements in learning spatial features, a persistent challenge remains to learn efficient representations that derive significant benefits from long-term temporal information in videos.

In this paper, we learn {\em online video representations} that incorporate multi-scale information on longer time horizons and design practical frameworks that achieve accuracy, efficiency and predictive power.  Analogous to the Gestalt principle of common fate \cite{Ellis_1938,Wertheimer_1938}, we hypothesize that temporal coherence by accounting for motion across frames allows learning powerful representations while achieving greater invariance to blur, lighting, pose and occlusions.  While our frameworks are applicable to diverse problems, we demonstrate them with the specific example of object detection in videos.

\begin{figure}[t]
\begin{center}
  \includegraphics[width=1.0\linewidth]{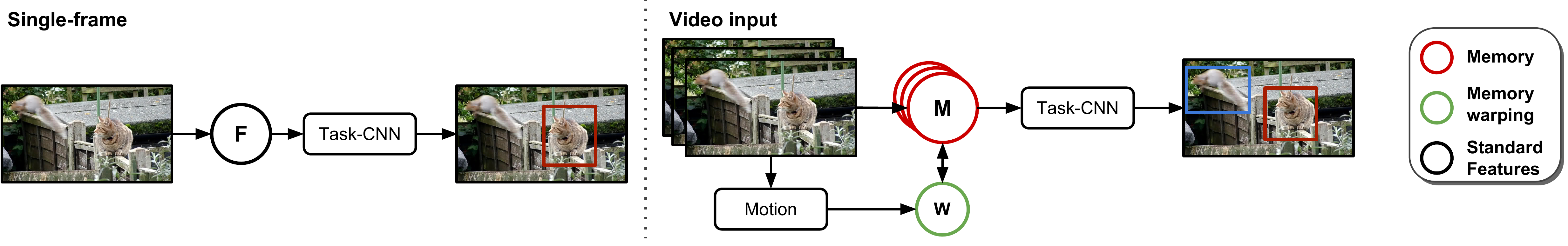}
\end{center}
\vspace{-0.7cm}
\caption{\small
  A schematic comparison between a typical per-frame method (left) and the proposed video representation learning approach (right) for {\em online} object detection in videos.  We propose a multi-scale memory that efficiently aggregates image evidence over longer time horizons and also accounts for camera and object motion by feature warping, which enables learning better representations that lead to higher accuracy.
}
\label{fig:teaser}
\end{figure}

In recent years, object detection in videos has attracted significant interest with benchmarks such as ImageNet VID~\cite{sam:Russakovsky15a} or Youtube-8M~\cite{sam:abuelhaija16a}.  A popular approach has been to use detected bounding boxes computed independently for each frame using a strong CNN-based model \cite{sam:Lin17b,sam:Lin17a,sam:Liu16a,sam:Ren15a} and to do temporal reasoning through tracking \cite{sam:Kang16a}, re-scoring detections \cite{sam:Feichtenhofer17c} and performing sequential non-maximum suppression \cite{Han_etal_2016}. While such methods improve over per-frame baselines, we explore the benefits of leveraging temporal information to learn better underlying feature representations. A few recent works temporally aggregate features to improve representation power~\cite{sam:Feichtenhofer17c,sam:Zhu17b}, but use a fixed set of nearby frames and do not maintain causality or efficiency.
In contrast, we propose a video representation that composes information across time in an {\em online} fashion (see Figure~\ref{fig:teaser}), which is not only faster, but also enables \emph{predictive} applications.

In particular, Section \ref{sec:memnet} proposes a novel network structure, termed {\em MemNet}, that holds a memory of the feature representation, which is updated at every frame based on image observations {\em solely from the past} and warped from one frame to the next to account for observer and scene motions.  We use a displacement field for warping similar to \cite{sam:Zhu17b}, but encoding memory allows retaining information from further in the past, while requiring only a single warp computation per-frame.
This is $2.3$ times faster than~\cite{sam:Zhu17b}, which requires as many warps as the number of temporally aggregated frames.\footnote{To put this speed-up in context for object detection, the improvement from Faster R-CNN \cite{sam:Ren15a} to R-FCN \cite{thvu:dai2016r} is about $2.5$ times.}
Further, in Section \ref{sec:clocknet}, we propose a hierarchical network structure, termed {\em ClockNet} due to similarity to \cite{sam:Koutnik14a,sam:Shelhamer16a}, which extends MemNet by operating at multiple temporal scales.  This allows efficiently leveraging information from even {\em longer temporal horizons}, which improves the representation power, as demonstrated by our experiments.

We apply our video representation to object detection in Section~\ref{sec:object_detection} and evaluate it on the ImageNet~VID~\cite{sam:Russakovsky15a} data set in Section~\ref{sec:exp_videodetection}. Our proposed architectures improve over per-frame baselines by up to 2.2\% in mean average precision (mAP).\footnote{To put this accuracy gain in context for object detection, the gains from hard example mining \cite{Wang_etal_2017} or hard positive generation \cite{Shrivastava_etal_2016} are around $2\%$ mAP.}  We achieve state-of-the-art results close to flow-guided feature aggregation (FGFA)~\cite{sam:Zhu17b} \emph{without having access to future frames} and with considerably lower runtime, while outperforming a causal variant of FGFA.

A key benefit of the online nature of our video representation is that it imparts predictive abilities, which enables novel applications.
First, in Section~\ref{sec:multigpu_detector}, we enhance the accuracy of an online real-time detector, by leveraging a stronger but less efficient detector in another thread.  While the strong detector lags due to higher latency, our memory warping enables {\em propagating and aligning} its representation with the real-time detector, boosting the accuracy of the latter by more than $10\%$ mAP, with no impact on speed or online operation (see Section~\ref{sec:exp_realtime_detector}).
This is non-trivial, since parallelizing standard detectors in an online setup is not straightforward.  Next, our predictive warping of video representations enables {\em anticipating} features in future frames, which allows solving visual tasks without actually observing future images.
Sections~\ref{sec:anticipating_objects} and \ref{sec:exp_featurepropagation} demonstrate this for the novel application of anticipating objects in future frames.

Finally, we note that our contributions are architecture-independent. The speed, accuracy and predictive benefits of our online representation are available for any detection method on video inputs.


\section{Related Work}
\label{sec:related_work}

Learning representations for videos has been a long-standing goal in computer vision and many directions have been explored.  Donahue~\etal~\cite{sam:Donahue15a} or Srivastava~\etal~\cite{sam:Srivastava15a} rely on recurrent neural networks (RNNs) like LSTMs~\cite{sam:Hochreiter97a} to propagate feature representation from still frames over time.  However, unlike our approach, features are propagated without explicit knowledge of motion in the scene, as we elaborate on in Section~\ref{sec:discussion}.  3D CNNs provide more freedom to learn motion-specific kernels (although motion is also not explicitly used) and were successfully used in~\cite{sam:Tran15a,sam:Yao15a} for tasks like video captioning or action recognition, but come with considerably more parameters to learn and typically more computational costs. Recent works have also considered unsupervised learning with pretext tasks \cite{Wang_Gupta_2015}, however, we focus on efficient learning of supervised representations.

Two-stream architectures like~\cite{sam:Simonyan14a,sam:Feichtenhofer16a,sam:Feichtenhofer17a} combine features extracted from both images and motion (optical flow) to boost the representational power.  While 3D CNNs learn motion-specific features implicitly, two-stream architectures explicitly take optical flow as input.  In both cases, however, this information is not used to transform features over time to compensate for observer and scene motions.

The recently proposed flow-guided feature aggregation framework (FGFA)~\cite{sam:Zhu17a,sam:Zhu17b}, on the other hand, explicitly warps convolutional feature maps between frames for better alignment when aggregating them.
The warping function is triggered by a learned displacement field initialized with FlowNet~\cite{thvu:dosovitskiy2015flownet}.  However, FGFA~\cite{sam:Zhu17b} uses a fixed temporal window of nearby frames, requires as many warp computations as the length of the window, and compromises causality, \ie, integrates features from future frames.  In contrast, we introduce a feature memory that is warped from one frame to another, saving computation time and allowing for a longer temporal horizon without looking into future frames.

Our temporal multi-scale variant, {\em ClockNet}, is inspired by~\cite{sam:Koutnik14a,sam:Shelhamer16a}.  While sharing the idea of efficiently extending the temporal horizon, we also motivate and experimentally demonstrate the multi-scale aspect of this architecture.  LSTMs~\cite{sam:Hochreiter97a} are also designed with a similar goal of a long-term representation for temporal data. The feature warping and propagation proposed in this work complements~\cite{sam:Koutnik14a,sam:Shelhamer16a,sam:Hochreiter97a} with an effective way of dealing with spatial memory maps that describe real video data where observer and the scene itself can be in motion.

We showcase our ideas for object detection in videos, which has recently attracted lots of interest, partly due to the ImageNet VID challenge~\cite{sam:Russakovsky15a}.  Besides~\cite{sam:Feichtenhofer17c,sam:Zhu17b} that work on the feature level, most recent approaches for detection in video leverage the temporal data on a higher level, \eg, by tracking objects or specifically designed post-processing mechanisms~\cite{sam:Han16a,sam:Kang16a}, which are orthogonal to our contributions.  Besides object detection, the proposed feature can also be used for other tasks operating on videos like semantic segmentation~\cite{sam:Kundu16a,sam:Shelhamer16a} or action recognition~\cite{sam:Feichtenhofer16a,sam:Feichtenhofer17a,sam:Simonyan14a}.  Visual anticipation has also been shown for segmentation \cite{Luc_etal_2017} and actions \cite{Vondrick_etal_2016}, but we differ in demonstrating for supervised object detection and in explicitly accounting for motion in the learned representation.


\begin{figure}[t]
\begin{center}
  \includegraphics[width=1.0\linewidth]{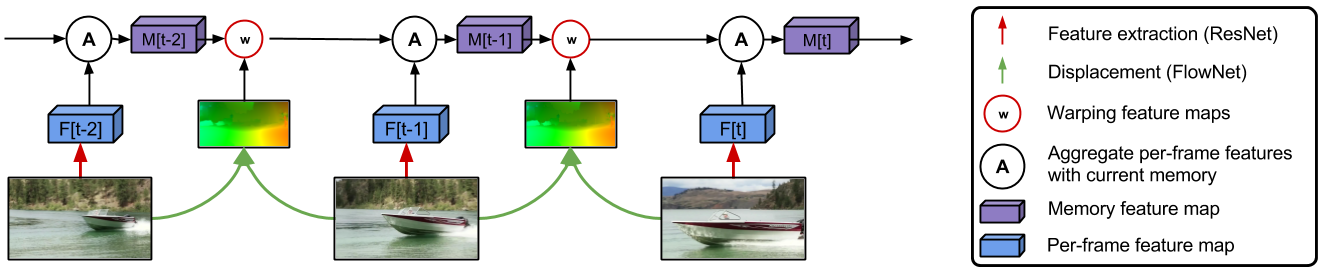}
\end{center}
\vspace{-0.6cm}
\caption{\small
  An illustration of the proposed \textbf{MemNet} running for three frames. At every time step, features from the current frame (blue) are aggregated with the memory of the previous frame (purple), either by simple averaging or a learned adaptive weighting. Then, the memory is warped via bilinear sampling based on a learned displacement field. The detection output in every frame is computed from the current memory.
}
\label{fig:memnet_structure}
\end{figure}

\section{Feature propagation via memory networks}
\label{sec:method}
The goal of this paper is to improve the feature representation for objects in videos, by leveraging temporal information and motion. Exploiting past frames can also help predictions in the current frame when occlusions or motion blur distorts image evidence, \cf~\cite{sam:Zhu17b}. We propose to continuously aggregate and update features over time to provide a stable and powerful representation of the scene captured by the video, which is illustrated in Figure~\ref{fig:memnet_structure}.

\subsection{Aggregating features over time}
\label{sec:memnet}
Given a single image $\image \in \realspace^{\imh \times \imw \times 3}$, a convolutional neural network (CNN) with parameters $\paramsFeat$ first extracts a feature map $\feat \in \realspace^{\feath \times \featw \times \featd}$, where $\featd$ is the number of feature maps and we typically have $\feath = \frac{1}{16}\imh$ and $\featw = \frac{1}{16}\imw$.  In the following, we show how these single image feature representations are effectively aggregated over time.  While we use a single feature map per image for ease of presentation, note that we can easily handle multiple feature maps at different resolutions to handle scale variations, which was shown to be useful in \cite{sam:Lin17b,sam:Yang16a}.

\paragraph{Tracking features over time:}
In every frame $t$, we hold a feature map $\memory_t \in \realspace^{\feath \times \featw \times \featd}$ that acts as a memory on the feature representation of the video.  Since the scene is dynamic and the camera is moving, the same objects will appear at different locations of the image plane in frames $t-1$ and $t$.
In order for the memory of the past frame $\memory_{t-1}$ to benefit detection in the current frame $t$, $\memory_{t-1}$ needs to be transformed according to the scene dynamics.
Similar to~\cite{sam:Zhu17b}, we use bilinear sampling to implement this transformation,
\begin{equation}
  \hat{\memory}_t = \bilinearSym(\memory_{t-1}, \flow_{(t,t-1)}) \;,
  \label{eq:warping_bilinear}
\end{equation}
where $\bilinearSym(\cdot)$ is the bilinear sampling function and $\flow_{(t,t-1)} \in \realspace^{\feath \times \featw \times 2}$ is a displacement (or flow) field between frames $t$ and $t-1$, which is estimated by a CNN with parameters $\paramsFlownet$.  This CNN is a pre-trained FlowNet~\cite{thvu:dosovitskiy2015flownet}, which takes images $\image_{t}$ and $\image_{t-1}$ as input and predicts the displacement, but we fine-tune the parameters $\paramsFlownet$ for the task at hand.  Note that for fast computation of the displacement field, we feed FlowNet with half-resolution images and up-scale the displacement field.  Also note that in the absence of ground truth data for the displacement field, this CNN predicts displacements suitable for the task at hand, which is demonstrated in Section~\ref{sec:exp_videodetection}.

\paragraph{Updating with image evidence:}
After having transformed the memory to the current frame $t$, \ie, $\hat{\memory}_{t}$, we need to aggregate the newly available image evidence $\feat_{t}$ extracted by the feature CNN into the memory,
\begin{equation}
  \memory_{t} = \aggregateSym(\hat{\memory}_{t}, \feat_{t}) \;,
  \label{eq:aggregate_generic}
\end{equation}
which defines one step of the proposed \emph{MemNet}.
We implement (and experimentally evaluate) two variants of the aggregation function $\aggregateSym(\cdot)$.
The first is a parameter-free combination that leads to exponential decay of memory over time,
\begin{equation}
  \aggregateSym(\hat{\memory}, \feat) := \frac{1}{2} (\memory + \feat) \;,
  \label{eq:aggregate_avg}
\end{equation}
and the second is a weighted combination of memory and image features,
\begin{equation}
  \aggregateSym(\hat{\memory}, \feat) := \wgtsMemory \cdot \memory + \wgtsFeat \cdot \feat \;,
  \label{eq:aggregate_wgts}
\end{equation}
with $\wgtsMemory,\wgtsFeat \in \realspace^{\feath \times \featw \times 1}$ and $\wgtsMemory + \wgtsFeat = \mat{1}$.
The weights are computed by a small CNN with parameters $\paramsWgtMemory$ and $\paramsWgtFeat$ operating on $\memory$ and $\feat$, respectively, and the constraint $\wgtsMemory + \wgtsFeat = \mat{1}$ is always satisfied by sending the concatenated output of the CNNs through a per-pixel softmax function.
The parameters of the weight-CNNs are automatically learned together with the rest of the network without any additional supervision.
In the first frame $t=1$, we simply assign the memory $\memory_1$ to be the feature representation of the image $\feat_1$.

\paragraph{Training MemNet:}
Training the proposed video representation requires a supervisory signal from a task module that is put on top of the memory features $\memory$.  In general, the task module can be anything, even an unsupervised task like predicting future frames~\cite{sam:Mathieu16a}.  In this work we explore object detection in videos, where the supervisory signal comes from a combination of object localization and classification loss functions, see Section~\ref{sec:object_detection}.

All parts of our representation can be trained end-to-end.  Since bilinear sampling and the grid generation of the warping module are both differentiable~\cite{sam:Jaderberg15a}, we can back-propagate gradients over time to previous frames, to the image feature extractor, as well as to the FlowNet generating the displacement fields.

While the network architecture in theory allows gradients to flow over the memory warping module to learn a good feature propagation, it also opens a shortcut for minimizing the loss because image evidence is available at every frame.  While for some tasks past information is truly essential for prediction in the present, for several tasks the image of the current frame already provides most evidence for a good prediction (or at least a signal to minimize the loss).  To encourage the network to learn a good feature propagation module, we randomly drop image evidence with probability $0.8$ at frame $t$, which we found to improve results by a few percentage points.

\begin{figure}[t]
\floatbox[{\capbeside\thisfloatsetup{capbesideposition={right,top},capbesidewidth=0.40\textwidth}}]{figure}[\FBwidth]
{\caption{\small Our \textbf{ClockNet} extends the MemNet by adding multiple time axes with increasing time scales to aggregate more information from further back in time.  Each additional time axis $k > 1$ skips $2^{k-1}-1$ frames.  We only illustrate two time scales to avoid clutter.}\label{fig:clocknet_structure}}
{\includegraphics[width=0.50\textwidth]{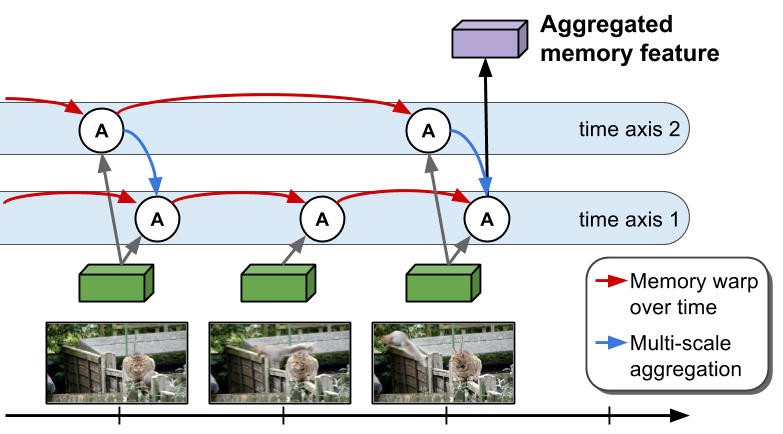}}
\vspace{-0.25cm}
\end{figure}

\subsection{Extending the temporal scale}
\label{sec:clocknet}
The basic MemNet operates on just a single temporal scale, which has limited capability to leverage information at a larger temporal horizon.
While, in theory, information from the whole video sequence is contained in the feature representation of the current frame $t$, this portion can be vanishingly small, particularly for the aggregation function relying on the exponential decay.

We thus propose to use a clock-work structure similar to~\cite{sam:Koutnik14a,sam:Shelhamer16a} that operates on multiple temporal scales, which we denote \emph{ClockNet} and illustrate in Figure~\ref{fig:clocknet_structure}.
Formally, instead of having a single memory feature map, we have $\numTimeScales$ memories $\memory_{t}^{k}$ at frame $t$ with $k \in \{1, \ldots, \numTimeScales\}$, each of them operating at different rates.
In our implementation, we update memory $\memory^{k}$ every $2^{k-1}$ frames with new image evidence, although other schedules are also possible.
Note that when $\numTimeScales = 1$, the basic \emph{MemNet} is obtained.

In order to exchange information across the different time scales $k$, we  aggregate all memory maps at a single frame $t$ by simply averaging them, \ie, $\memory_t = \frac{1}{\numTimeScales} \sum_{k=1}^{\numTimeScales} \memory_t^k$.
As with the feature map aggregation in the basic \emph{MemNet}, different strategies for combining feature maps are possible.  We chose the simpler parameter-free averaging, as a more complex learning-based weighting scheme did not show any performance gains.  The aggregated memory $\memory_t$ can then be used as input to any task-specific modules.

\begin{figure}[t]
\floatbox[{\capbeside\thisfloatsetup{capbesideposition={right,top},capbesidewidth=0.35\textwidth}}]{figure}[\FBwidth]
{\caption{\small Relation between standard convolutional recurrent neural networks (left) and the proposed video representation (right).}\label{fig:relation_to_conv_rnns}}
{\includegraphics[width=0.63\textwidth]{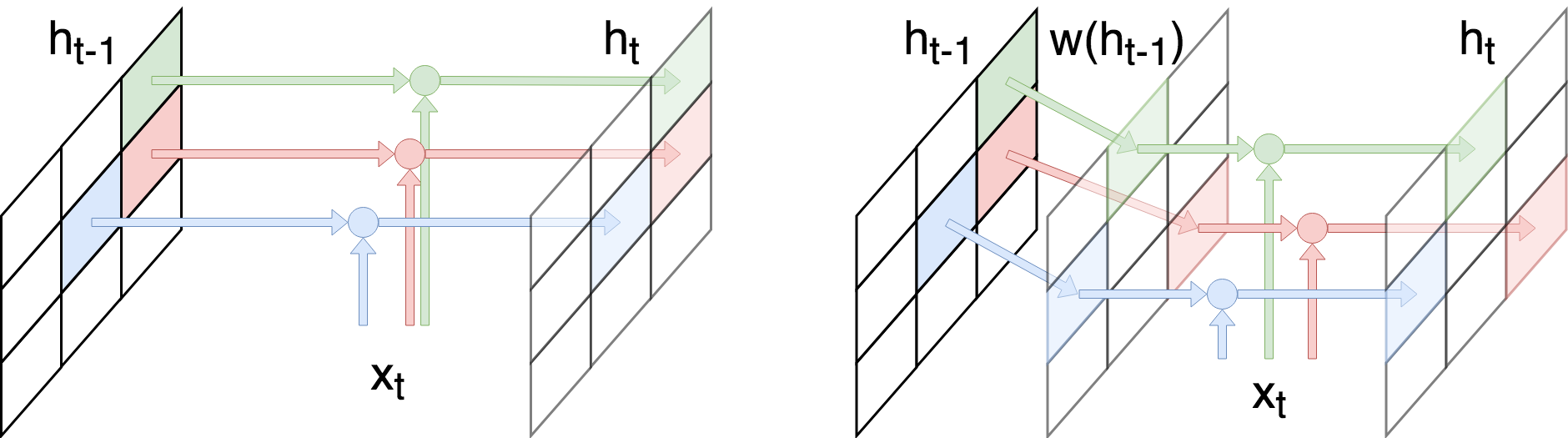}}
\vspace{-0.25cm}
\end{figure}

\subsection{Discussion}
\label{sec:discussion}
Our proposed video representations have a simple and intuitive structure, can be trained end-to-end and fulfill the basic requirements for a fast and causal system that can be applied to videos in any real-world application.  In contrast to FGFA~\cite{sam:Zhu17b}, the proposed model does not look at future frames and is also not limited to a specific temporal horizon in the past, rather can carry information from the whole (past) sequence in its memory.  An even longer temporal horizon is utilized by the ClockNet architecture.

There also exists a relation to convolutional recurrent neural networks (cRNN) \cite{sam:Pinheiro14a,sam:Liang15a}, however, with one crucial difference.  While cRNNs keep their hidden memory fixed across spatial dimensions ($h_t = \textrm{RNN}(h_{t-1},x_t)$), our model enables the memory to be spatially aligned with observer and scene motion in the actual video content ($h_t = \textrm{RNN}(\textrm{warp}(h_{t-1}, \flow_{t,t-1}), x_t)$), see Figure~\ref{fig:relation_to_conv_rnns}.  While our aggregation function $\aggregateSym(\cdot)$ for new input and previous hidden states is simple, we did not observe any improvements for our particular applications with more complex architectures like LSTM~\cite{sam:Hochreiter97a} or GRU~\cite{sam:Cho14a}.


\section{Detection, Propagation and Anticipation in Videos}
\label{sec:applications}
To demonstrate the benefits of propagating features over time with the proposed \emph{MemNet} and \emph{ClockNet}, we show its impact for three practical applications.

\subsection{Object detection in videos}
\label{sec:object_detection}
While we can use the proposed memory features $\memory$ from Section~\ref{sec:method} for any downstream task, in this paper, we focus on object detection in videos.  Modern object detectors like Faster-RCNN~\cite{sam:Ren15a}, R-FCN~\cite{thvu:dai2016r}, SSD~\cite{sam:Liu16a} or RetinaNet~\cite{sam:Lin17a} all have a similar high-level structure in the sense that they all rely on a convolutional neural network to extract features $\feat$ from a single image.  The detection-specific modules applied on top of $\feat$ define the differences between the detectors, \eg, proposal-based~\cite{sam:Ren15a,thvu:dai2016r} or proposal-free~\cite{sam:Liu16a,sam:Lin17a}, making $\feat$ an interface between one generic module and detection-specific modules.  Our proposed {\em MemNet} and {\em ClockNet} operate on $\feat$ and compute a novel feature representation $\memory$, making our video representation applicable to all of these detectors. In this paper, we pick R-FCN~\cite{thvu:dai2016r} because it has shown a good trade-off between accuracy and speed, with a publicly available code base.

Given a representation $\memory_t$ of a video sequence at frame $t$, the object detector first computes object proposals with a region proposal network (RPN) as proposed in~\cite{sam:Ren15a}.  Object proposals define potential locations of objects of interest (independent of the actual category) and reduce the search space for the final classification stage.  Each proposal is then classified into one of $\ncls$ categories and the corresponding proposal location is further refined.  In contrast to Faster-RCNN~\cite{sam:Ren15a}, the per-proposal computation costs in R-FCN are minimal by using position-sensitive ROI pooling.  This special type of ROI pooling is applied on the output of the region classification network (RCN).

\begin{figure}[t]
  \centering
  \includegraphics[width=0.55\textwidth]{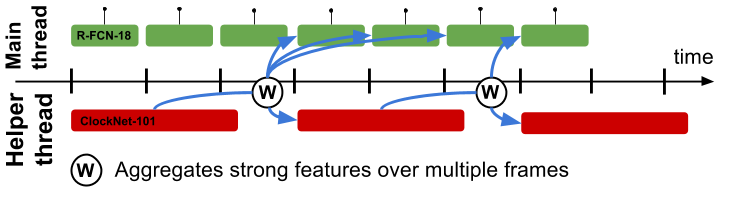}
  \vspace{-0.35cm}
  \caption{\small In a multi-GPU setup, a fast but weak object detector (R-FCN-18, green blocks) leverages the features of a strong but slow object detector (ClockNet-101, red blocks), see Section~\ref{sec:multigpu_detector}. The width of blocks represent computation time. At a frame $t$, strong features from ClockNet-101 are only available from $t-\Delta$, but efficiently warped into frame $t$ with our propagation module. The warped features boost the representational power of R-FCN-18 significantly, without increasing latency of the real-time system.}
  \label{fig:twothreaded_detector}
\end{figure}

\subsection{Real-time detection by propagating strong features}
\label{sec:multigpu_detector}
Assuming an input stream capturing images at 20 frames-per-second (FPS), we ideally want an object detector that can process one image in less than 50 ms to avoid latency in the output.  One easy option to speed-up a modern object detector is to use a more light-weight feature extraction CNN, \eg, by using ResNet-18 instead of ResNet-101.  Note that this is a viable option for any detection framework, \eg, Faster-RCNN~\cite{sam:Ren15a}, R-FCN~\cite{thvu:dai2016r}, YOLO~\cite{sam:Redmon16a,sam:Redmon17a} or SSD~\cite{sam:Liu16a}.  However,  accuracy will decrease.  Here, we explore another option to speed-up a modern object detector.  Instead of using a single model, we demonstrate how to exploit two models with complementary properties running simultaneously (but asynchronously) on two threads (two GPUs) to achieve both speed and accuracy, using our feature propagation.

We run a fast detector, R-FCN-18 (\ie, R-FCN with ResNet-18~\cite{thvu:he2016deep}) in one thread and a slower but also stronger detector, ClockNet-101, in the other thread.  R-FCN-18 runs at the required frame rate and can provide output for every frame, however at a lower quality than ClockNet-101 could do if no time requirements existed.  The main problem with the strong object detector is that it will always have some delay (or latency) $\Delta$ to produce an output.  If $\Delta$ is too large for a practical system, the strong detector is not usable.  It is important to note that achieving a speed-up with two GPUs is not trivial in a real-time setting.  For the offline case it is easy to distribute computation of different images on multiple GPUs.  However, this is not an option for streaming data.  In Section~\ref{sec:exp_realtime_detector}, we still empirically compare with two alternative baselines that also leverage two GPUs.

With our design, on the other hand, we can still leverage the strong features by making up for the delay via feature propagation.  We compute the displacement field between frame $t+\Delta$ and $t$ and warp the strong features $\memory_{t}^{101}$ into the current frame $t+\Delta$, where the fast object detector has already computed features $\feat_{t+\Delta}^{\textrm{18}}$, see Figure~\ref{fig:twothreaded_detector}.  We boost the representational power of R-FCN-18 by combining the feature maps.  Again, we take the average of both features (the dimensionality is the same), but more advanced aggregation schemes are possible.  We experimentally evaluate this application in Section~\ref{sec:exp_realtime_detector}.

\subsection{Anticipating features}
\label{sec:anticipating_objects}
Another application of the proposed feature propagation is future prediction or anticipation.  Features from the current frame $t$ are propagated to a future frame $t+\Delta$, where the task network is applied to make predictions. In the previous application of Section~\ref{sec:multigpu_detector}, we utilize feature propagation over several frames, but the displacement fields are still computed from image evidence, similar to~\cite{sam:Zhu17a}. For a true visual anticipation, however, future images are not available.

We propose to extrapolate the displacement fields into future frames and use them to propagate the feature (or memory) maps. For demonstration, we use a simple extrapolation technique.  Given the two previous displacement fields $\flow_{t-1,t-2}$ and $\flow_{t,t-1}$, we compute the difference of aligned displacement vectors (with bilinear sampling), which gives us the acceleration of pixels.  We then employ a simple constant acceleration motion model to each displacement vector and extrapolate for one or multiple frames. Obviously, this extrapolation technique has limitations but it is sufficient for our demonstrations of feature anticipation. We analyze the quality of the anticipated features in Section~\ref{sec:exp_featurepropagation} by measuring the object detection quality in future frames.


\section{Experiments}
\label{sec:experiments}
Our experimental evaluation focuses on the performance of our feature propagation and aggregation methods for object detection in videos.  In Sections~\ref{sec:exp_videodetection}, \ref{sec:exp_featurepropagation} and \ref{sec:exp_realtime_detector}, we evaluate the performance of the proposed \emph{MemNet} and \emph{ClockNet} on the three applications introduced in Section~\ref{sec:applications}, respectively.

\paragraph{Dataset:}
All our experiments are conducted on the ImageNet VID data set~\cite{sam:Russakovsky15a}, which is most suitable for object detection in videos and has been the testbed for most recent approaches for this task~\cite{sam:Feichtenhofer17c,sam:Kang16a,sam:Zhu17a,sam:Zhu17b}.  ImageNet VID is a large scale data set consisting of $5344$ video clips, captured at frames rates between $25$ and $30$ FPS and divided into training, validation and testing sets with $3862$, $555$ and $937$ clips respectively.  Each clip is fully annotated with bounding box tracks of $30$ different object classes.

\paragraph{Implementation details:}
We use the ResNet-101 architecture~\cite{thvu:he2016deep} as the basic feature extractor in all experiments with the exception of some ablation studies.  In particular, we use \textit{\`{a} trous} convolutions as in~\cite{thvu:chen2016deeplab,sam:Zhu17b} to increase the feature resolution.  Similar to~\cite{sam:Zhu17b} the extracted features are passed through a $3\times3$ convolutional layer and a non-linear activation (ReLU~\cite{thvu:nair2010rectified}) before we provide them as inputs to {\em MemNet} and {\em ClockNet}.  To estimate displacement fields we use FlowNet~\cite{thvu:dosovitskiy2015flownet}.  All the parameters are jointly fine-tuned end-to-end.
In general, we closely follow the experimental setup of~\cite{sam:Zhu17b} using their publicly available MXNet~\cite{thvu:MxNet} implementation.  All models are trained for $2$ epochs with an initial learning rate of $0.001$, which is decreased by a factor of $10$ after $\frac{4}{3}$ epochs.  We train our models on the same mix of ImageNet DET and ImageNet VID training sets as in~\cite{sam:Zhu17b}.  All experiments, including runtime measurement, are done on an NVIDIA TITAN Xp.  For training, we use 4 GPUs.

\subsection{Object detection in videos}
\label{sec:exp_videodetection}
Object detection in videos aims at localizing objects in every video frame, \ie, estimating bounding boxes around objects associated with a confidence score.

\paragraph{Evaluation metrics:}
We measure detection performance as mean average precision (mAP) over all object classes, where we additionally differentiate between fast, medium and slowly moving objects using the subsets of videos introduced in~\cite{sam:Zhu17b}.  We also measure the average runtime per frame in milliseconds (ms) for each model (using the same framework and GPU setup).

\paragraph{Baselines:}
We compare the proposed {\em MemNet} and {\em ClockNet} models with several baselines.  The first one is the per-frame object detector itself (R-FCN~\cite{thvu:dai2016r,sam:Zhu17b}) that does not exploit temporal information.  The second baseline is FGFA~\cite{sam:Zhu17b}, which, for every frame, aggregates features from nearby frames both in the past and the future. Obviously, this makes FGFA a non-causal system not applicable to real-time tasks.  Note that these two baselines represent two extremes of using temporal information, with R-FCN not exploiting the video at all and FGFA looking not only into the past but also into future frames.  For a more fair comparison, we thus created a causal variant of FGFA (Cau. FGFA) that aggregates information only from nearby features in the past but not from future frames.  While FGFA can only operate in the off-line setting where future frames are accessible, causal FGFA is an on-line detector, making it the most comparable baseline to our proposed models.

\begin{figure}[t]
\begin{floatrow}
\ffigbox[\FBwidth]{%
  \includegraphics[width=0.4\textwidth,trim={2 2 2 2},clip]{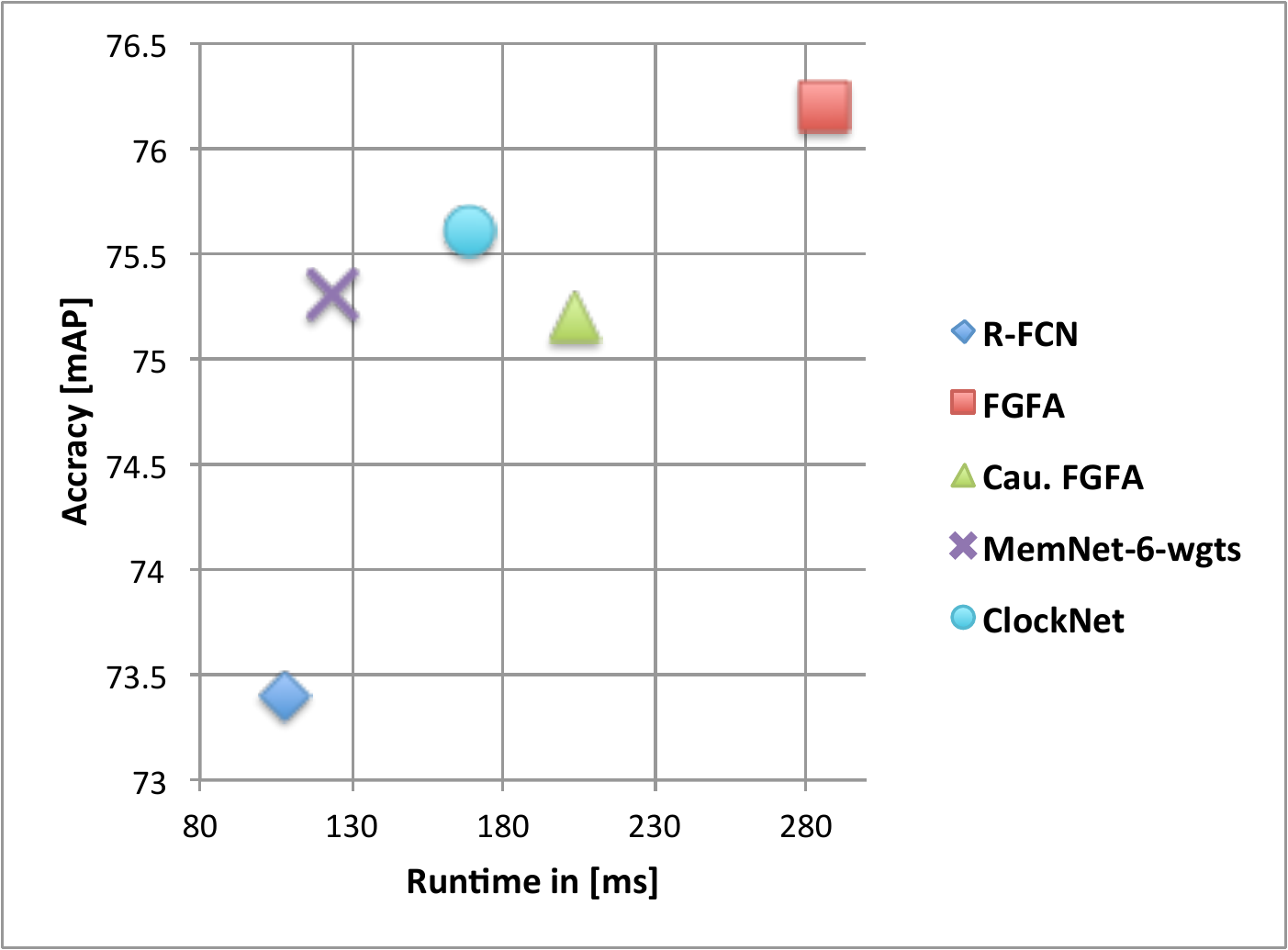}%
}{%
  \caption{\small Accuracy and runtime trade-off of various methods.}\label{fig:acc_runtime_tradeoff}%
}
\capbtabbox[\Xhsize]{%
  \scriptsize\begin{tabular}{l c c c c c}
  \toprule
  Method       & \parbox{0.6cm}{mAP}& \parbox{0.8cm}{\centering mAP\\(fast)} & \parbox{0.8cm}{\centering mAP\\(med)} & \parbox{0.8cm}{\centering mAP\\(slow)} & ms \\
  \toprule
  R-FCN~\cite{thvu:dai2016r} & 73.4 & 51.4 & 71.6 & 82.4 & 108 \\
  FGFA~\cite{sam:Zhu17b}  & 76.2 & 56.0 & 75.2 & 83.8 & 286 \\
  \midrule
  FGFA (half) & 66.7 & 42.2 & 66.1 & 77.5 & 152 \\
  Cau. FGFA    & 75.2 & 53.9 & \textbf{74.1} & 83.8 & 204 \\
  Cau. FGFA*  & 66.0 & 40.0    & 65.3    & 79.3 & 181 \\
  MemNet-3     & 74.3 & 51.9 & 72.5 & 83.6 & 122 \\
  MemNet-6     & 75.1 & 53.8 & 73.5 & 83.3 & 122 \\
  MemNet-6-wgts  & 75.3 & 51.8 & 73.6 & 83.8 & 124 \\
  MemNet-6-strd-4 & 74.4 & 51.7 & 72.4 & \textbf{84.2} & 122 \\
  MemNet-6-strd-8 & 74.2 & 51.6 & 72.6 & 82.7 & 122 \\
	ClockNet     & \textbf{75.6} & \textbf{55.4} & 73.7 & 83.4 & 169 \\
  \bottomrule
\end{tabular}


}{%
  \vspace{-0.25cm}
  \caption{\small Detection performance and runtime on ImageNet-VID validation of different methods.}%
  \label{tbl:imagenetvid_mainres}
}
\end{floatrow}
\end{figure}

\begin{figure}[t]
  \begin{center}
    \begin{subfigure}[t]{0.48\textwidth}\centering
            \includegraphics[width=0.95\textwidth,height=3.4cm]{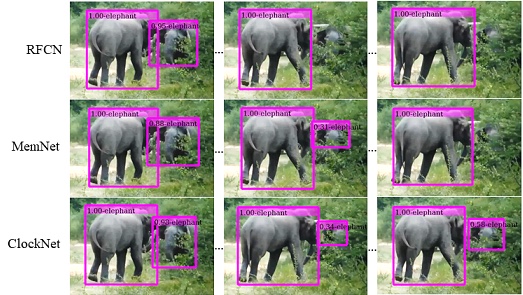}
      \caption{}
      \label{fig:memory_visualization_a}
    \end{subfigure}
    \quad
    \begin{subfigure}[t]{0.48\textwidth}\centering
      \includegraphics[trim={65, 0, 0, 0},clip,width=0.95\textwidth,height=3.4cm]{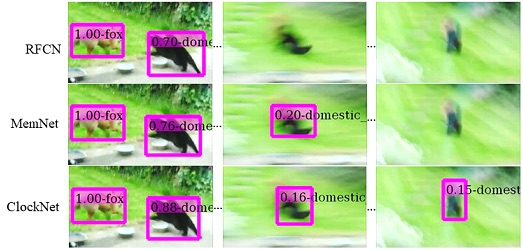}
      \caption{}
      \label{fig:memory_visualization_b}
    \end{subfigure}
    \\
    \begin{subfigure}[t]{0.48\textwidth}\centering
      \includegraphics[width=0.95\textwidth,height=2.8cm]{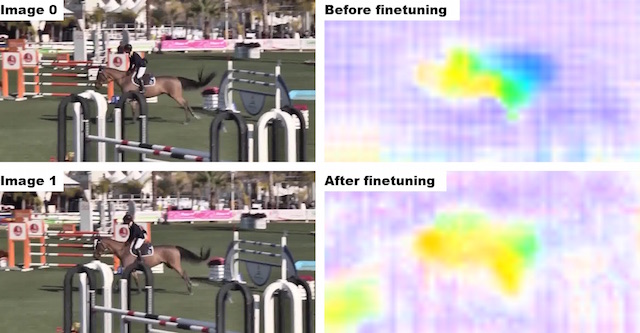}
      \caption{}
      \label{fig:flow_fields}
    \end{subfigure}
    \quad
    \begin{subfigure}[t]{0.48\textwidth}\centering
      \includegraphics[trim=1 1 1 1,clip,width=0.95\textwidth,height=2.8cm]{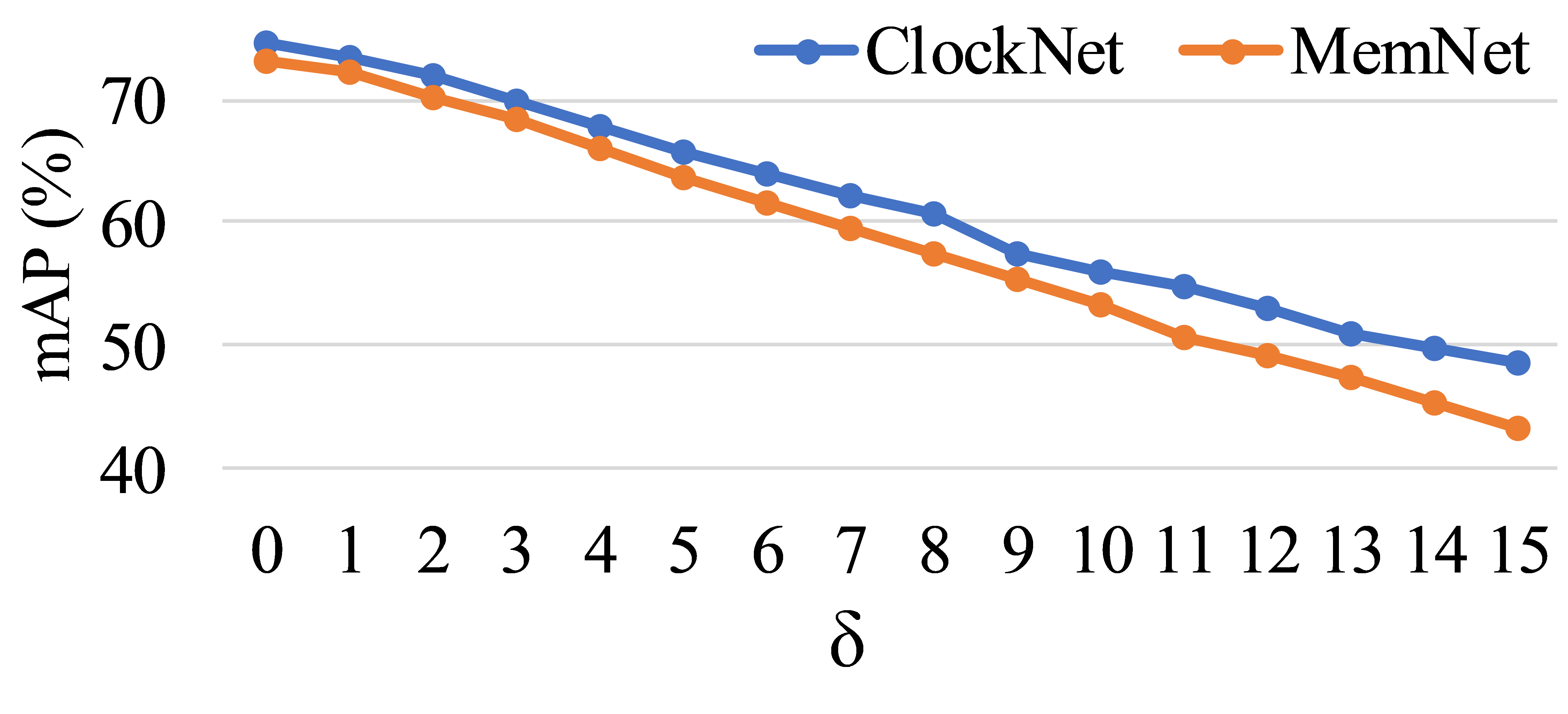}
      \caption{}
      \label{fig:exp_feature_propagation}
    \end{subfigure}
  \end{center}
\vspace{-0.1cm}
\caption{\small
  (a-b) Qualitative examples of our memory models and the RFCN baseline.
  (c) The flow fields before and after fine-tuning demonstrate the impact of jointly fine-tuning the FlowNet with the object detector.
  (d) Mean AP of MemNet and ClockNet with respect to different propagation lengths $\delta$.
}
\vspace{-0.5cm}
\label{fig:combo_qualitresults_propagation}
\end{figure}

\paragraph{Main results:}
Table~\ref{tbl:imagenetvid_mainres} summarizes our quantitative results and Figures~\ref{fig:memory_visualization_a}-\ref{fig:memory_visualization_b} give qualitative examples.  We can first see that all models leveraging temporal data improve over the per-frame baseline R-FCN~\cite{thvu:dai2016r}.  Looking into both past and future frames, as FGFA~\cite{sam:Zhu17b} does, gives the best overall results, but  comes at a considerable runtime cost and, more importantly, is a non-causal system.  Among all causal systems leveraging data only from the past (eight bottom models in Table~\ref{tbl:imagenetvid_mainres}), the proposed {\em ClockNet} gives the best results overall and is particularly strong for fast moving objects.  Its mAP value is only 0.6 percentage points behind FGFA without having access to future frames.

Looking at the running times, we see that the proposed {\em MemNet} is clearly the fastest, except for the still-frame baseline, and {\em ClockNet} already ranks second.  Both proposed models are faster than causal FGFA and consequently, FGFA~\cite{sam:Zhu17b}.  Built upon the detection framework of R-FCN, all those models have the same computational complexity for feature extraction, proposals generation and classification.  Therefore, the causes of speed difference are the numbers of flow computations $N_{F}$ and feature warps $N_{W}$.  For FGFA and Causal FGFA, $N_F$ and $N_W$ equal the number of frames within aggregation range, i.e. 20 and 10, respectively.  The {\em ClockNet} reported in Table~\ref{tbl:imagenetvid_mainres} requires $N_F=3$ and $N_W=3$, corresponding to its $3$ temporal scales, and {\em MemNet} has a speed advantage because it only needs $N_F=1$ and $N_W=1$ to process an incoming frame.

We also note that the computation time of Causal FGFA can be reduced by aggregating displacement fields in an online manner, see Cau. FGFA* in Table~\ref{tbl:imagenetvid_mainres}, thus reducing $N_{F}$ to 1 as in MemNet.  While the runtime is reduced, the error accumulation in online aggregation of displacement fields leads to a significant accuracy drop.  To highlight the reduction in runtime from MemNet and ClockNet, we show another trivial way to speed-up FGFA by halving the image resolution, FGFA (half).  However, this variant also leads to a large performance drop.

Overall, accuracy and runtime demonstrate the advantages of our memory propagation mechanism for object detection in videos, see Figure~\ref{fig:acc_runtime_tradeoff}.  Moreover, our memory-based architectures have the additional benefit (over FGFA) of being able to propagate features into future frames, which we analyze in Section~\ref{sec:exp_featurepropagation} and demonstrate in a practical application in Section~\ref{sec:exp_realtime_detector}.

\paragraph{Ablation studies:}
First, we investigate the importance of the length of the sequences used for training {\em MemNet}.  In the testing phase there is no limitation on the sequence length, but GPU memory is a hard constraint during training as gradients need to flow back through the whole sequence.  We can observe in Table~\ref{tbl:imagenetvid_mainres} that a longer temporal window for training {\em MemNet} ({\em MemNet-6} vs. {\em MemNet-3}) is indeed beneficial.
Second, by comparing different aggregation schemes (Equations~\eqref{eq:aggregate_avg} and \eqref{eq:aggregate_wgts}), we can observe that learning an additional weighting gives a small improvement with only a tiny increase in computational costs.
We also tried LSTM~\cite{sam:Hochreiter97a} as aggregation scheme, but did not observe any improvements.
Finally, we analyze the importance of multiple temporal scales in {\em ClockNet}, which operates on three time axes $k=[1,3,4]$, corresponding to temporal strides of 1, 4 and 8, respectively.  To emphasize the benefit of feature aggregation across multiple temporal scales, we compare {\em ClockNet} to {\em MemNet} trained with longer strides ({\em MemNet-6-strd-4} and {\em MemNet-6-strd-8}), given it access to larger temporal horizon during training.  We can see from Table~\ref{tbl:imagenetvid_mainres} that both baselines perform worse than {\em MemNet-6} (with temporal stride of 1), which illustrates the importance of the information provided by actual neighboring frames.  Therefore, the success of {\em ClockNet} provides a signal for the benefits of a temporal multi-scale architecture, where different temporal scales are complementary to each other.  More experiments on various details of our methods can be found in the supplemental material.

\paragraph{Fine-tuning FlowNet:}
Finally, we want to understand the effect of fine-tuning FlowNet during training the video representation for the detection task.
Similar to~\cite{sam:Zhu17b}, we observe a performance drop in mAP if FlowNet is not fine-tuned.
Given that displacement fields after fine-tuning are apparently better for the detection task, we visualize the difference for one example in Figure~\ref{fig:flow_fields}.  One difference that we can observe is that the object tends to move as a whole and ignores the motion of individual parts.

\subsection{Propagating and anticipating features}
\label{sec:exp_featurepropagation}
In the previous experiment, the appearance feature of the current frame is always available and the main purpose was to analyze the influence of additional information from the past.
In this experiment, we want to give more insights into the quality of the feature propagation and anticipation performance, which has several applications as discussed in Section~\ref{sec:applications}.

\paragraph{Propagation:} In this experiment, we provide image information up to frame $t-\delta$ and then propagate up to frame $t$.  Note that displacement fields are still available but only for warping the memory; no image evidence in form of appearance features is available to the detector after frame $t-\delta$.
The propagated features are then used to compute the detections.  We compare our model with a baseline that takes the detected bounding boxes at frame $t-\delta$ and propagates them with the computed displacement fields to frame $t$.  The mean displacement inside a bounding box serves as the translational vector.
Table~\ref{tbl:imagenetvid_featprop} shows the impact of skipping the image features for different amounts of frames for MemNet, ClockNet and the box-propagating baselines.  In general, all models gracefully degrade performance with larger $\delta$, but at the same time reduce running time.  It is evident that feature propagation outperforms propagation on the bounding box level, particularly for ClockNet.  From Figure~\ref{fig:exp_feature_propagation}, we can see that the performance gap between MemNet and ClockNet increases when $\delta$ gets larger, demonstrating the impact of multiple temporal scales and the extend time horizon of ClockNet.

\begin{table}[t]\scriptsize
\begin{floatrow}
\capbtabbox[\FBwidth]{%
  \begin{tabular*}{0.45\textwidth}{l @{\extracolsep{\fill}} ccc|ccc}
  \toprule
  Methods  & \multicolumn{3}{c|}{mAP} & \multicolumn{3}{c}{FPS} \\
  $\delta$ & 0 & 4 & 8 & 0 & 4 & 8 \\
  \midrule
  \multicolumn{7}{c}{\textit{Box - Propagation}} \\
  MemNet          & 75.1 & 64.6 & 55.4     & 8.2 & 14.7  & 16.9 \\
  ClockNet        & 75.6 & 64.9 & 56.2     & 5.9 & 12.5  & 15.3 \\
  \midrule
  \multicolumn{7}{c}{\textit{Feature - Propagation}} \\
  MemNet          & 75.1 & 68.9 & 56.1     & 8.2 & 14   & 16 \\
  ClockNet        & 75.6 & 70.9 & 62.3     & 5.9 & 6.6  & 8.3\\
  \midrule
  \multicolumn{7}{c}{\textit{Feature - Anticipation}} \\
  MemNet          & -    & 68.8 & 55.9     & - & 5.1 & 8.7 \\
  ClockNet        & -    & 67.0 & 57.3     & - & 2.5 & 3.4 \\
  \bottomrule
\end{tabular*}

%
}{%
  \vspace{-0.25cm}
  \caption{\small Runtime in fps and accuracy in mAP of MemNet and ClockNet for feature propagation and anticipation.}\label{tbl:imagenetvid_featprop}%
}
\capbtabbox[\Xhsize]{%
  \begin{tabular}{l c c c}
  \toprule
    Method                           & \#G  & mAP   & FPS  \\
  \midrule
  R-FCN-18                         & 1       & 60.2  & 14.3 \\
  ClockNet-101                     & 1       & 75.6  &  5.9 \\
  ClockNet-101-FeatProp          & 1       & 70.9  &  6.6 \\
  \midrule
  R-FCN-18 (split)                 & 2       & 21.7     & 28.6    \\
  ClockNet-101 (split)             & 2       & 28.9     & 11.8    \\
  ClockNet-101-FeatProp (split)        & 2       & 23.3     & 13.2    \\
  R-FCN-18+ClockNet-101-BoxProp	& 2		 & 66.9		& 14.3	  \\
  \midrule
  R-FCN-18+ClockNet-101-FeatProp       & 2       & 71.9  & 14.3 \\
  \bottomrule
\end{tabular}

%
}{%
  \vspace{-0.25cm}
  \caption{\small Accuracy and runtime of our two-threaded detection setup.  The number of GPUs utilized is denoted as \#G.}\label{tbl:imagenetvid_twothreads}%
}
\end{floatrow}
\end{table}

\paragraph{Anticipation:}
We next evaluate feature anticipation as discussed in Section~\ref{sec:anticipating_objects}, which differs to the previous experiment since no image information is available at all for frames after $t-\delta$.  This requires us to extrapolate the displacement field as described in Section~\ref{sec:anticipating_objects}.  Comparing the anticipation results with the propagation model in Table~\ref{tbl:imagenetvid_featprop}, we can see that our flow extrapolation strategy works well for MemNet which has short temporal stride, although the runtime speed drops due to our current non-optimized implementation of flow extrapolation.  Reversely, on ClockNet memories, feature anticipation performs worse than feature propagation.  One explanation is that quality of the extrapolated flows is heavily degraded with the long temporal strides of ClockNet.

\subsection{Fast detection by propagating strong features}
\label{sec:exp_realtime_detector}
In this section we analyze the application described in Section~\ref{sec:multigpu_detector} and Figure~\ref{fig:twothreaded_detector}.  A fast but weak object detector is running in the main thread providing detection output at every frame, while leveraging features from a strong but slow feature extractor.  In order to use the strong features, they have to be propagated to the current frame, \ie, aligned over time, to compensate for the delay $\Delta$.

We use R-FCN based on the ResNet-18 architecture as the fast main-thread detector, which runs at $14.3$ FPS.  The helper thread runs {\em ClockNet} based on ResNet-101 (same as in Table~\ref{tbl:imagenetvid_mainres}), which is able to propagate features over time as demonstrated in Section~\ref{sec:exp_featurepropagation}.  In practice, ClockNet-101 runs $2.4$ times slower than R-FCN-18.  To compensate for additional overhead (\eg, feature propagation), we update ClockNet-101 with image evidence once every $4$ frames (``ClockNet-101-FeatProp'').  For training the fast detector, \ie, R-FCN-18, we only leverage fixed features coming from the second thread but do not fine-tune ClockNet-101.

\paragraph{Quantitative results:}
Table~\ref{tbl:imagenetvid_twothreads} shows the results of this experiment.  As expected, R-FCN-18 is the fastest model but also gives the worst accuracy.  ClockNet-101 is the model shown in Section~\ref{sec:exp_videodetection} and can be considered an upper bound in terms of accuracy but is very slow compared to R-FCN-18.  ClockNet-101-FeatProp is the model running in the helper thread which receives image evidence every 4 frames and uses propagation to make predictions in other frames.  The performance drop compared to the upper bound is not much, as also seen in Section~\ref{sec:exp_featurepropagation}, but the runtime is still high.
However, when following the design proposed in Section~\ref{sec:multigpu_detector}, \ie, feeding R-FCN-18 with propagated strong features from ClockNet-101-FeatProp, we observe a significant 10\% mAP boost compared to R-FCN-18 which is even close to the ClockNet-101 upper bound, while maintaining the low runtime of R-FCN-18.

\paragraph{Comparison to two-thread baselines:}
Recall that achieving speed-up with two GPUs in a real-time (data-streaming) setting is non-trivial as parallelizing frames over multiple GPUs is not possible.
We still evaluate alternative baselines that leverage two GPUs.  First, a trivial two times speed-up can be achieved by splitting the image into two halves, running the detector individually and merging the detections before NMS, denoted ``(split)'' in Table~\ref{tbl:imagenetvid_twothreads}.  While this gives the expected speed-up, the performance drop is significant, which can be explained by the fact that objects in Imagenet-VID are mostly centered, thus effectively truncating them.
The second baseline is more evolved and similar to our proposed design.  However, instead of propagating features, (delayed) detections from the strong model are propagated to align with the faster detector (``BoxProp''), as in Section~\ref{sec:exp_featurepropagation}.  As in the previous experiment, we observe that feature-level propagation is superior to propagating bounding boxes.

\section{Conclusions}
\label{sec:conclusions}
Our work proposes a long-term online video representation that effectively leverages past information.  We rely on a memory that is updated regularly with image evidence and is warped over time for a proper spatio-temporal alignment of features.  We also extend this representation to multiple temporal scales, thus aggregating information from even farther in the past.  Our experiments illustrate the benefits of each component, as well as a demonstration of two practical applications: feature anticipation and a real-time multi-threaded object detector.

For future works, we plan to investigate the effectiveness of our learned video representation for action recognition and segmentation in videos. We also plan to further explore the utility of feature anticipation and postulate that causally learned representations like ours have inherent advantages for visual anticipation.


\bibliographystyle{splncs}
\bibliography{myshortstrings,mem_rfcn_biblist}

\pagestyle{headings}
\mainmatter

\title{Supplemental Material:\\Memory Warps for Learning Long-Term \\ Online Video Representations}

\author{Tuan-Hung Vu$^{1}$ $\;$ Wongun Choi$^2$ $\;$ Samuel Schulter$^3$ $\;$ Manmohan Chandraker$^{3,4}$}
\institute{INRIA/ENS WILLOW$^1$ $\quad$ ISEE$^2$ $\quad$ NEC-Labs$^3$ $\quad$ UC San Diego$^4$}

\maketitle

The supplemental material contains the following items:
\begin{itemize}
\item \textbf{Section~\ref{sec:results_per_category}:} Detailed per-category results of our main evaluation
\item \textbf{Section~\ref{sec:ablation}:} Additional ablation studies of the proposed MemNet
\item \textbf{Section~\ref{sec:qualitative_results}:} Additional qualitative results of the proposed video detectors
\item \textbf{Section~\ref{sec:anticipation}:} Additional details on the feature anticipation experiment
\end{itemize}

\section{Per-category results on ImageNet-VID data set}
\label{sec:results_per_category}
\begin{figure}[h!]
	\begin{center}
    \begin{subfigure}[t]{1.0\columnwidth}
      \centering
      \includegraphics[trim=5 5 5 5,clip,width=1.0\columnwidth]{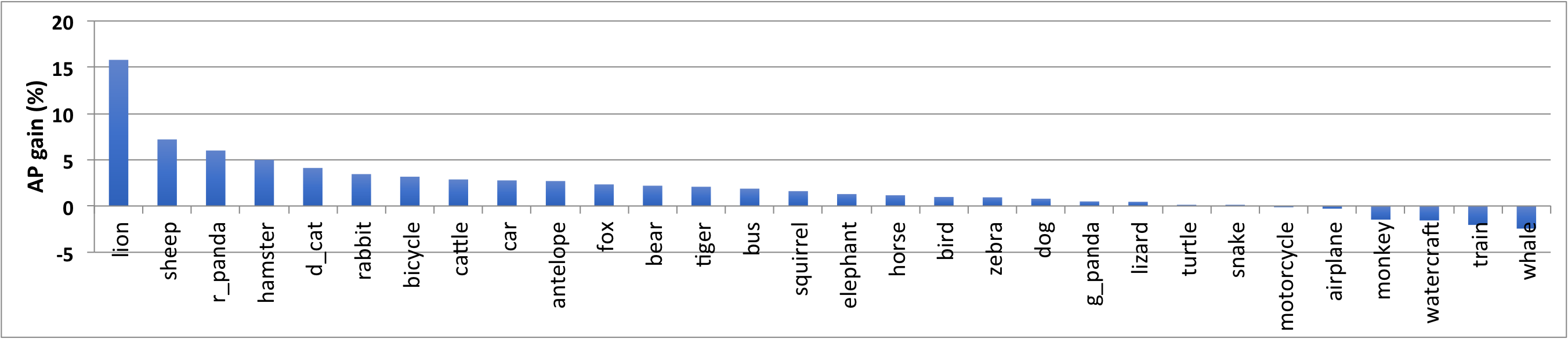}
      \caption{All}
    \end{subfigure}

    \begin{subfigure}[t]{1.0\columnwidth}
      \centering
      \includegraphics[trim=5 5 5 5,clip,width=1.0\columnwidth]{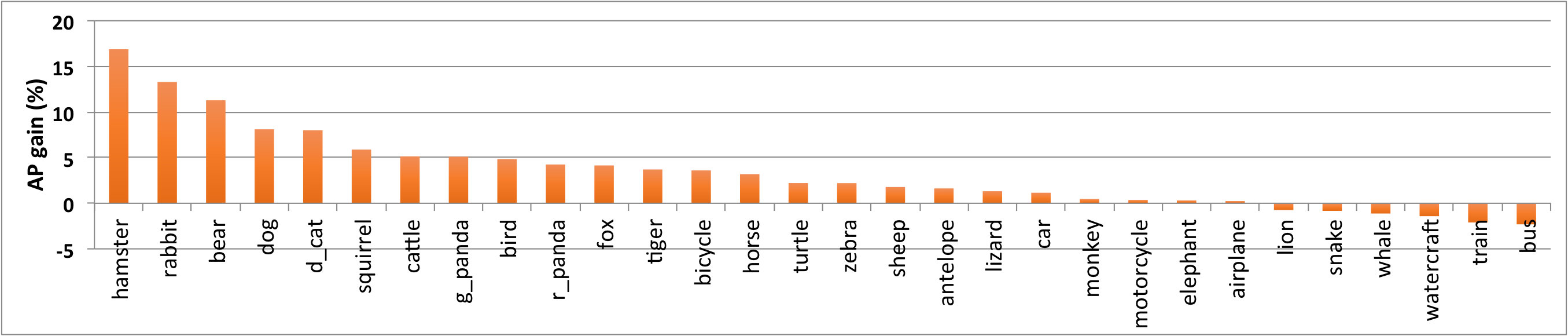}
      \caption{Fast}
    \end{subfigure}

    \begin{subfigure}[t]{1.0\columnwidth}
      \centering
      \includegraphics[trim=5 5 5 5,clip,width=1.0\columnwidth]{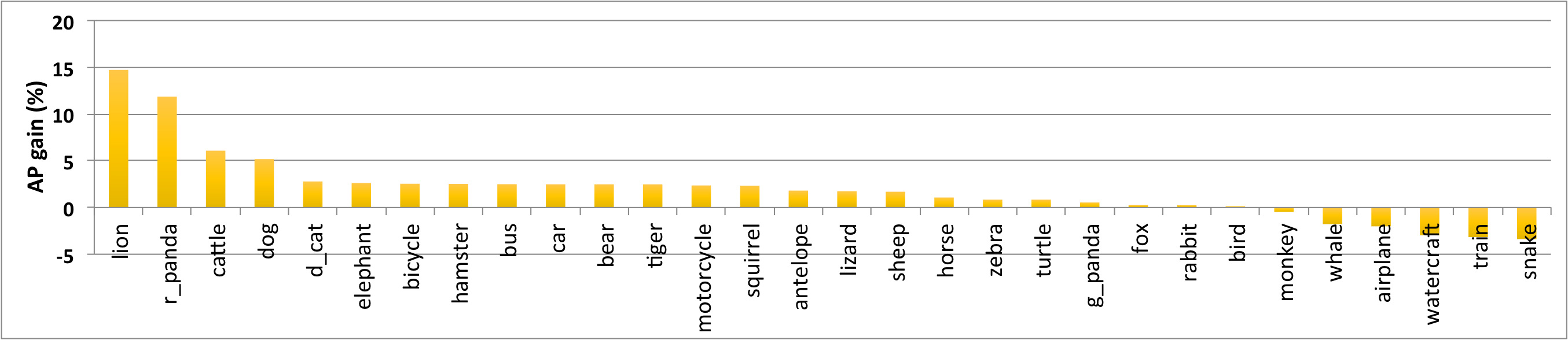}
      \caption{Medium}
    \end{subfigure}

    \begin{subfigure}[t]{1.0\columnwidth}
      \centering
      \includegraphics[trim=5 5 5 5,clip,width=1.0\columnwidth]{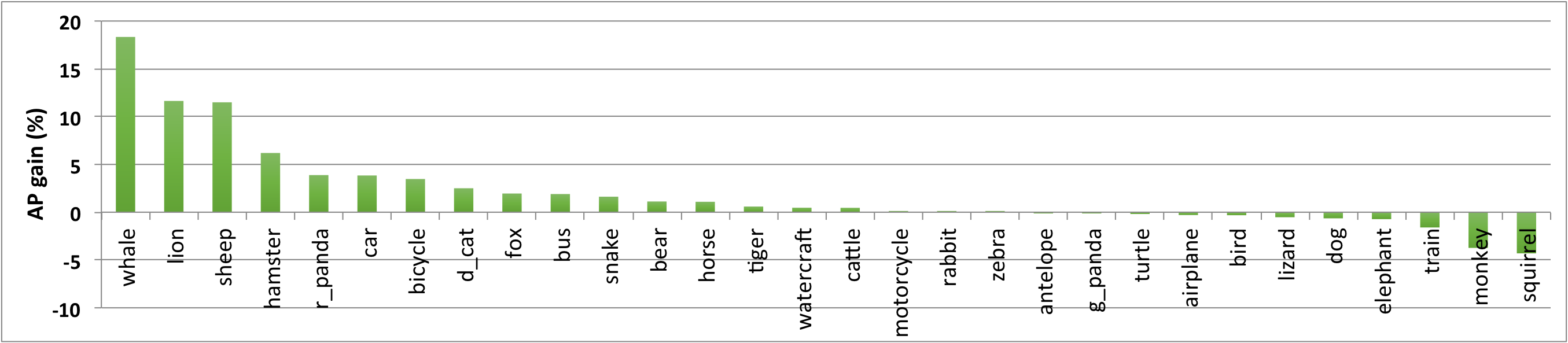}
      \caption{Slow}
    \end{subfigure}
	\end{center}
	\caption{Per-category performance gain of ClockNet over R-FCN~\cite{thvu:dai2016r} on ImageNet-VID validation set (a) as well as the fast, medium and slow speed splits (b-d).}
	\label{fig:AP_all}
\end{figure}

The main results in Table 1 of the main paper show the mean of the average precision value (AP) across all 30 categories of the ImageNet-VID data set.  In Table~\ref{tbl:imagenetVID_details}, we provide the per-category results for the main methods: R-FCN~\cite{thvu:dai2016r}, FGFA~\cite{sam:Zhu17b}, Causal FGFA, MemNet-6 and ClockNet.  To better illustrate the performance gain of leveraging temporal data, we show the per-category AP gain of ClockNet over R-FCN~\cite{thvu:dai2016r} in Figure~\ref{fig:AP_all}, where each sub-figure shows different subsets of the data set corresponding to the columns of Table 1 in the main paper: \textbf{``all''} corresponds to all frames of the data set. \textbf{``fast, medium, slow''} correspond to subsets of the frames where objects move fast, normal (medium fast) and slow.  These subsets have been defined in~\cite{sam:Zhu17b}.

\begin{table}[h!]\centering
	\begin{tabular}{lc@{\;\;\;}c@{\;\;\;}c@{\;\;\;}c@{\;\;\;}c@{\;\;\;}c@{\;\;\;}c@{\;\;\;}c@{\;\;\;}c@{\;\;\;}c@{\;\;\;}c}
  \toprule
  &\rotatebox{90}{\small airplane} &\rotatebox{90}{\small antelope} & \rotatebox{90}{\small bear} &	\rotatebox{90}{\small bicycle} &	\rotatebox{90}{\small bird} &	\rotatebox{90}{\small bus} &	\rotatebox{90}{\small car} &	\rotatebox{90}{\small cattle} &	\rotatebox{90}{\small d\_cat} &	\rotatebox{90}{\small dog} \\
  & \includegraphics[width=0.4cm]{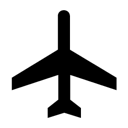}& \includegraphics[width=0.4cm]{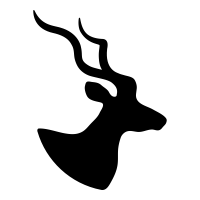}& \includegraphics[width=0.4cm]{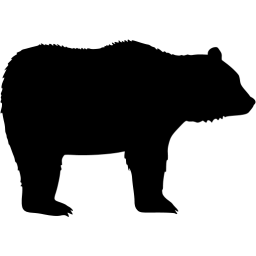}& \includegraphics[width=0.4cm]{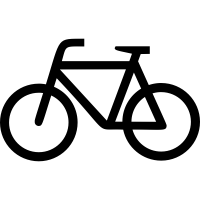}& \includegraphics[width=0.4cm]{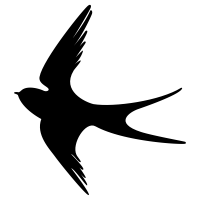}& \includegraphics[width=0.4cm]{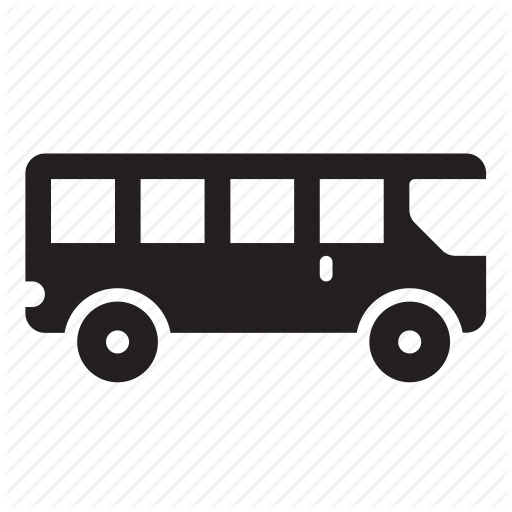}& \includegraphics[width=0.4cm]{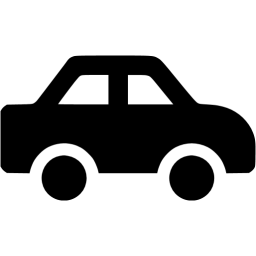}& \includegraphics[width=0.4cm]{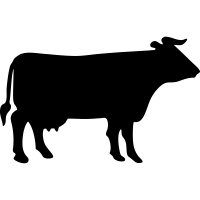}& \includegraphics[width=0.4cm]{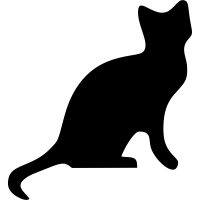} & \includegraphics[width=0.4cm]{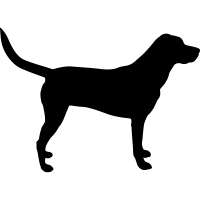} & \\
  \midrule
  R-FCN~\cite{thvu:dai2016r}  & 89.3 &	77.9 &	82.2 &	68.5 &	71.9 &	77.0 &	56.5 &	67.5 &	69.6 &	82.0 & \\
  FGFA~\cite{sam:Zhu17b}      & 88.5 &	80.1 &	83.6 &	71.6 &	73.4 &	79.3 &	61.0 &	71.5 &	73.5 &	83.6 & \\
  Causal-FGFA                 & 87.1 &	78.7 &	85.8 &	71.8 &	74.8 &	80.2 &	58.6 &	68.2 &	73.1 &	83.1 & \\
  MemNet-6                    & 89.0 &	80.9 &	82.2 &	68.3 &	73.2 &	77.1 &	57.9 &	69.9 &	69.8 &	83.7 & \\
  ClockNet                    & 89.0 &	80.6 &	84.4 &	71.6 &	72.9 &	78.9 &	59.3 &	70.4 &	73.7 &	82.8 & \\
  \vspace{-0.4cm}\\
\toprule
  &	\rotatebox{90}{\small elephant} &	\rotatebox{90}{\small fox} &	\rotatebox{90}{\small g\_panda}&	\rotatebox{90}{\small hamster} &	\rotatebox{90}{\small horse} &	\rotatebox{90}{\small lion} & \rotatebox{90}{\small lizard}&	\rotatebox{90}{\small monkey} &	\rotatebox{90}{\small motorbike} &	\rotatebox{90}{\small r\_panda} \\
  & \includegraphics[width=0.4cm]{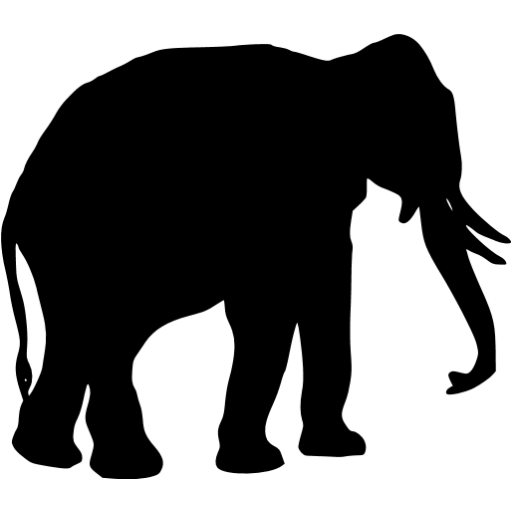}& \includegraphics[width=0.4cm]{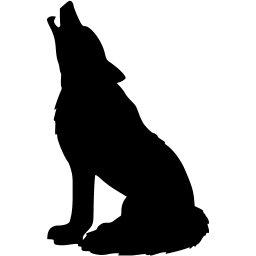}& \includegraphics[width=0.4cm]{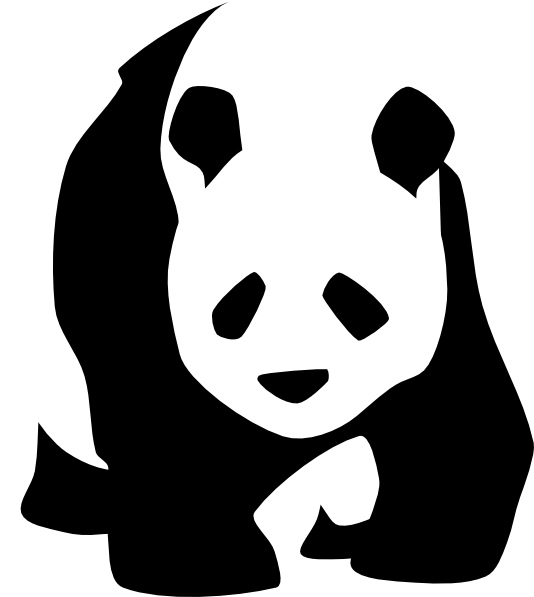}& \includegraphics[width=0.4cm]{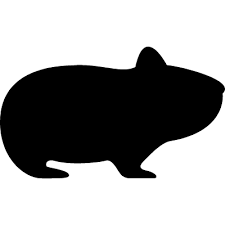}& \includegraphics[width=0.4cm]{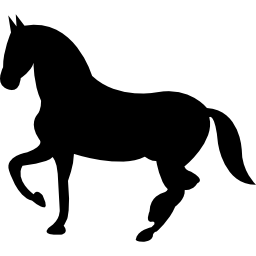}& \includegraphics[width=0.4cm]{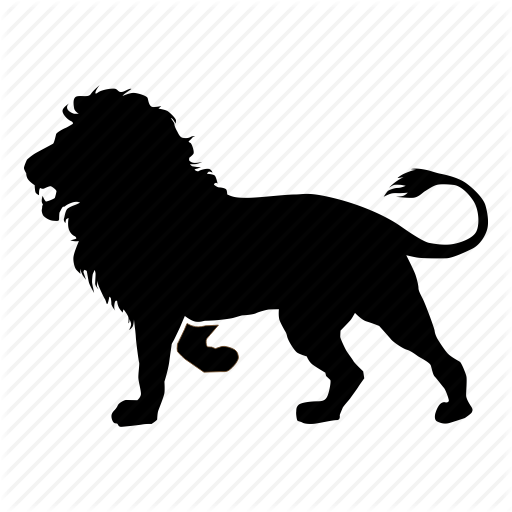} & \includegraphics[width=0.4cm]{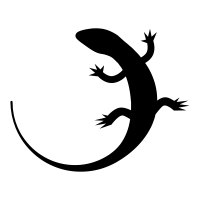}& \includegraphics[width=0.4cm]{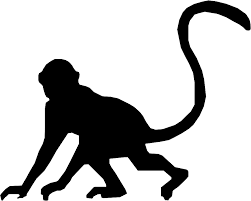}& \includegraphics[width=0.4cm]{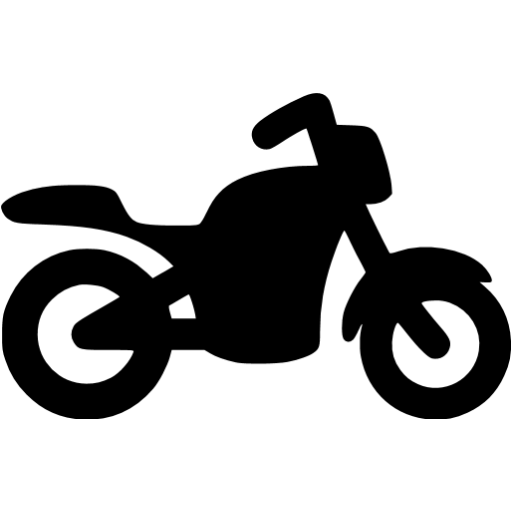}& \includegraphics[width=0.4cm]{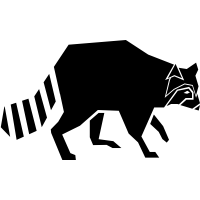} & \\
  R-FCN~\cite{thvu:dai2016r} & 75.3 &	87.2 &	78.7 &	89.6 &	69.9 &	41.5 & 77.9 &	51.6 &	81.2 &	63.1 & \\
  FGFA~\cite{sam:Zhu17b}     & 77.0 &	89.9 &	80.9 &	94.0 &	70.8 &	55.9 & 79.0 &	50.9 &	84.2 &	63.4 & \\
  Causal-FGFA                & 77.2 &	89.6 &	82.2 &	94.1 &	76.1 &	48.3 & 78.7 &	50.6 &	82.2 &	61.2 & \\
  MemNet-6                   & 75.1 &	89.5 &	80.1 &	88.6 &	72.1 &	64.8 & 80.0 &	52.8 &	83.5 &	64.0 & \\
  ClockNet                   & 76.6 &	89.5 &	79.2 &	94.6 &	71.1 &	57.3 & 78.3 &	50.1 &	81.1 &	69.0 & \\
  \toprule
  &	\rotatebox{90}{\small rabbit} &	\rotatebox{90}{\small sheep} &	\rotatebox{90}{\small snake} &	\rotatebox{90}{\small squirrel} &	\rotatebox{90}{\small tiger} & \rotatebox{90}{\small train} &	\rotatebox{90}{\small turtle} &	\rotatebox{90}{\small watercraft}	& \rotatebox{90}{\small whale} &	\rotatebox{90}{\small zebra} & mAP \\
  & \includegraphics[width=0.4cm]{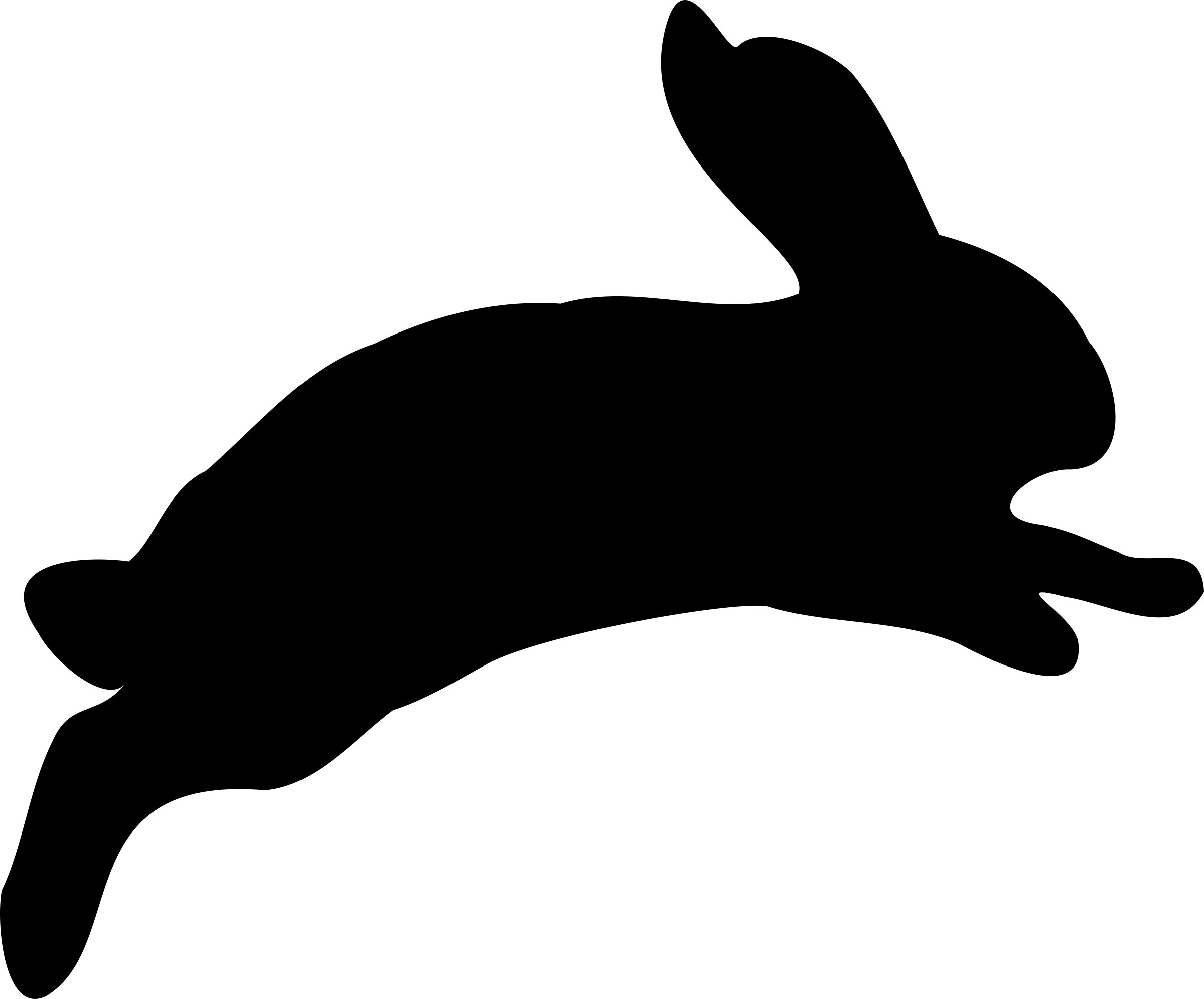} & \includegraphics[width=0.4cm]{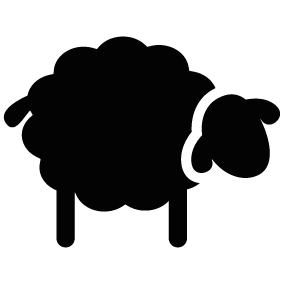}& \includegraphics[width=0.4cm]{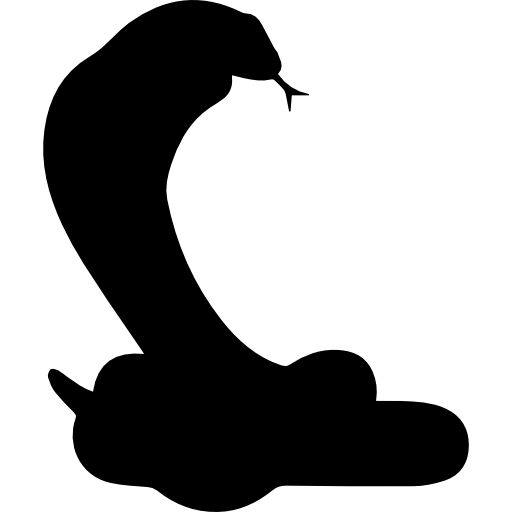}& \includegraphics[width=0.4cm]{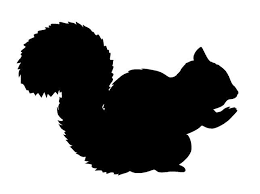}& \includegraphics[width=0.4cm]{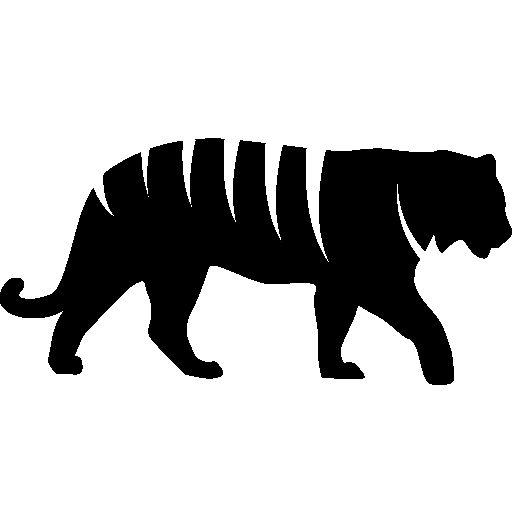}& \includegraphics[width=0.4cm]{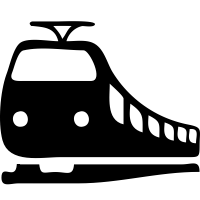}& \includegraphics[width=0.4cm]{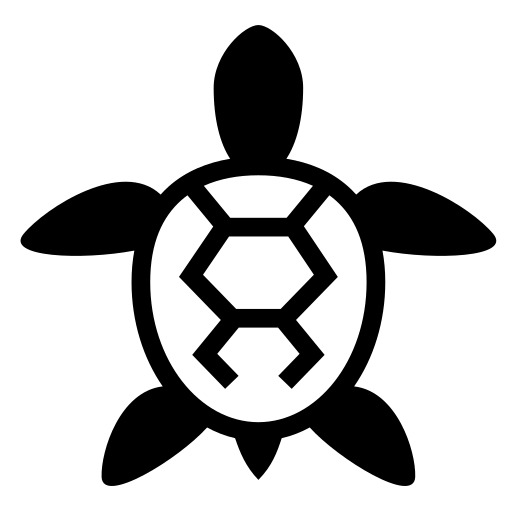}& \includegraphics[width=0.4cm]{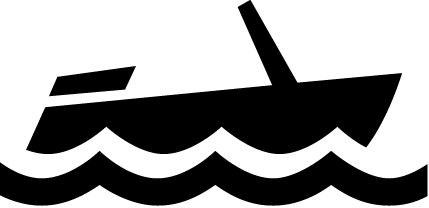}& \includegraphics[width=0.4cm]{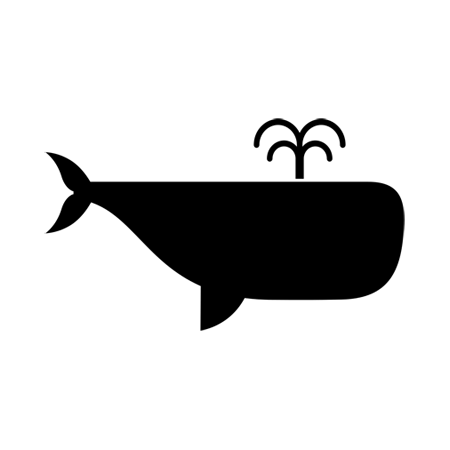}& \includegraphics[width=0.4cm]{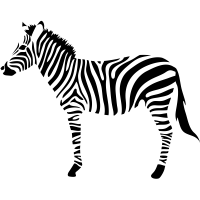} & \\
  \midrule
  R-FCN~\cite{thvu:dai2016r} & 80.7 &	51.0 &	75.5 &	55.4 &	90.1 &	82.6 &	80.0 &	68.2 &	74.5 &	91.0 & 73.6 \\
  FGFA~\cite{sam:Zhu17b}     & 86.9 &	62.2 &	74.8 &	60.9 &	91.7 &	81.3 &	81.0 &	67.4 &	75.3 &	91.2 & 76.2 \\
  Causal-FGFA                & 78.5 &	57.8 &	76.7 &	58.8 &	90.4 &	81.6 &	81.9 &	67.5 &	71.9 &	89.8 & 75.2 \\
  MemNet-6                   & 78.1 &	57.8 &	75.7 &	58.3 &	90.8 &	84.1 &	78.5 &	64.0 &	72.4 &	90.8 & 75.1 \\
  ClockNet                   & 84.2 &	58.2 &	75.5 &	57.0 &	92.1 &	80.6 &	80.1 &	66.7 &	72.0 &	91.9 & 75.6 \\
  \bottomrule
\end{tabular}


	\caption{Per-category AP ($\%$) on ImageNet VID validation set.}
	\label{tbl:imagenetVID_details}
\end{table}

\section{Additional results of ablation studies}
\label{sec:ablation}
This section provides (i) a comparison between different aggregation schemes for MemNet, (ii) an analysis of the underlying feature representation our memory module is applied to and (iii) additional qualitative examples of the learned displacement fields.  Note that we conduct ablation studies of MemNet with only half resolution images for computational reasons, which explains lower accuracies for both, baselines and our proposed methods.

First, we complement the evaluation of different aggregation schemes between memory and features from new image evidence in Section 5.1 of the main paper.  Table~\ref{tbl:diff_mem_agg} compares simple averaging and the weighting described in the main paper with adaptive weighting from~\cite{sam:Zhu17b} and the recently proposed AdaScan~\cite{thvu:kar2016adascan}.  As already discussed in the main paper, it is interesting to see that simple averaging gives on par results with more complex weighting schemes.

Second, we apply the proposed MemNet on different levels of a ResNet-50 feature extractor: ``conv3'', ``conv4'', ``conv4+conv5'', and ``conv5''.  For this particular application of object detection in videos, one can observe from Table~\ref{tbl:diff_rep_lvl} that ``conv5'' gives the best results overall.  Other applications may require different setups.

\begin{figure}[t]
	\begin{center}
		\includegraphics[width=0.49\linewidth]{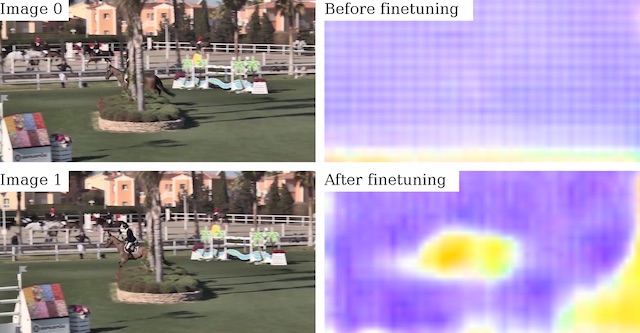} \includegraphics[width=0.49\linewidth]{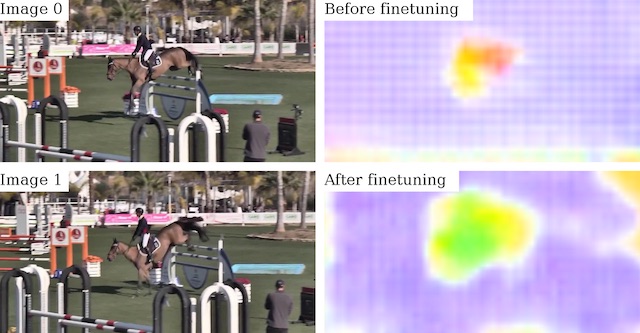} \\
		\includegraphics[width=0.49\linewidth]{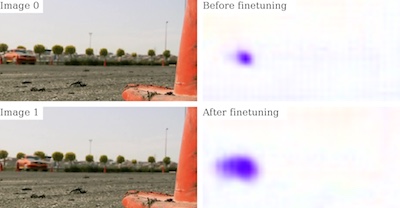} \includegraphics[width=0.49\linewidth]{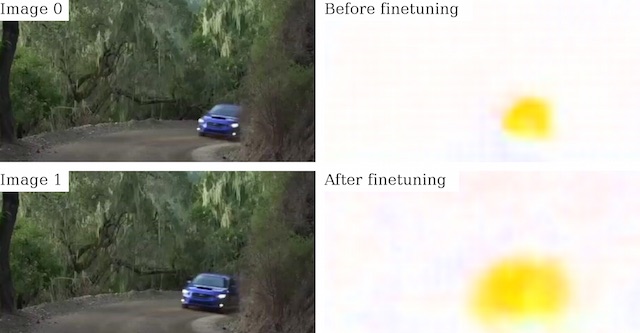} \\
		\includegraphics[width=0.49\linewidth]{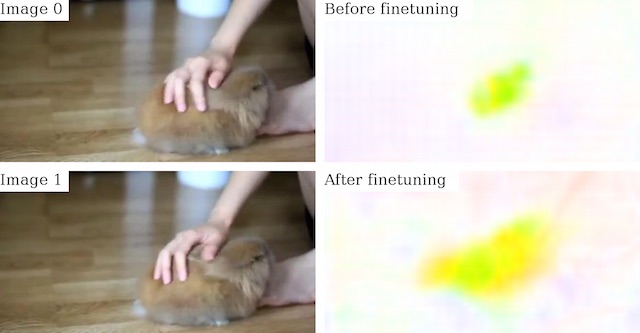} \includegraphics[width=0.49\linewidth]{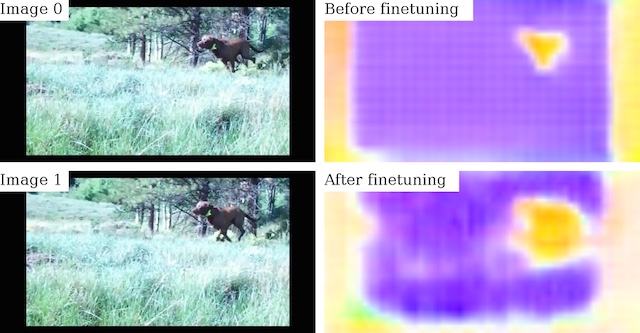} \\
		\includegraphics[width=0.49\linewidth]{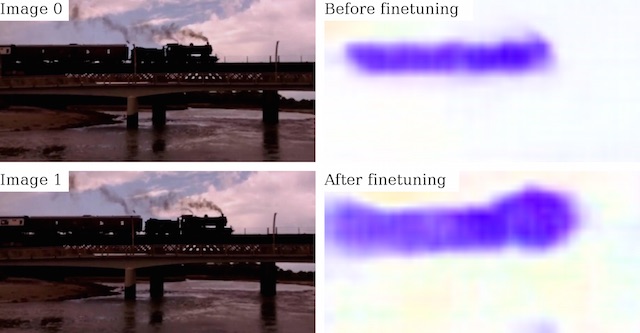} \includegraphics[width=0.49\linewidth]{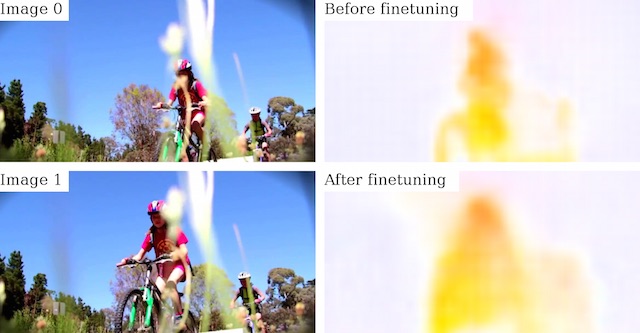} \\
	\end{center}
	\caption{We illustrate the impact of jointly fine-tuning the FlowNet with the object detector.  For each example consists of 4 images: the 2 input images (``Image0, Image1'') and the displacement field before and after fine-tuning.  The figure shows 2 columns and 4 rows of examples.
	}
	\label{fig:flow_fields_supmat}
\end{figure}

Finally, we also provide more examples of the learned displacement fields in Figure~\ref{fig:flow_fields_supmat}.  As in the example in the main paper, we can see that after fine-tuning FlowNet, objects tend to move as a whole, while different motion of individual parts is suppressed.  Moreover, we also see some examples where the initial FlowNet only provides motion for parts of an objects but after fine-tuning the whole object is covered, thus refining the flow.

\begin{table}[h]\centering
  \begin{subtable}{1.0\textwidth}\centering
    \begin{tabular}{l c c c c}
      \toprule
      & \parbox{2.5cm}{\centering Average} & \parbox{2.5cm}{\centering Weighting} & \parbox{2.5cm}{\centering Adaptive Weighting~\cite{sam:Zhu17b}}  & \parbox{2.5cm}{\centering AdaScan~\cite{thvu:kar2016adascan}} \\ 
      \midrule
      mAP & 66.4 & 66.4 & 66.4 & 66.2 \\ 
      \bottomrule
    \end{tabular}
    \caption{}
    \label{tbl:diff_mem_agg}
  \end{subtable}\vspace{0.5cm}
  \\
  \begin{subtable}{1.0\textwidth}\centering
    \begin{tabular}{l c c c c }
      \toprule
      & \parbox{2.5cm}{\centering conv3} & \parbox{2.5cm}{\centering conv4} & \parbox{2.5cm}{\centering conv4+conv5} & \parbox{2.5cm}{\centering conv5} \\
      \midrule
      mAP & 66 & 66 & 66.2 & 66.4 \\
      \bottomrule
    \end{tabular}
    \caption{}
    \label{tbl:diff_rep_lvl}
  \end{subtable}
  \caption{Detection performance of MemNet in ablation studies on ImageNet-VID validation set.  (a) shows the impact of different aggregation schemes and (b) shows different levels of the feature representation where the memory module is employed.}
  \label{tbl:ablation_studies}
\end{table}

\section{Qualitative results}
\label{sec:qualitative_results}
\begin{figure}[h!]\centering
	\begin{tabular}{@{}l@{\,\,\,}l@{\,}l@{\,}l@{}}
    \textbf{RFCN}&\includegraphics[trim=0 40 0 70,clip,width=0.24\linewidth]{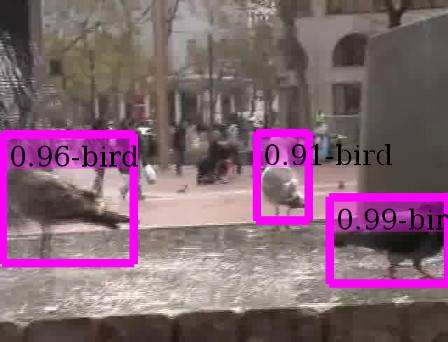}&\includegraphics[trim=0 40 0 70,clip,width=0.24\linewidth]{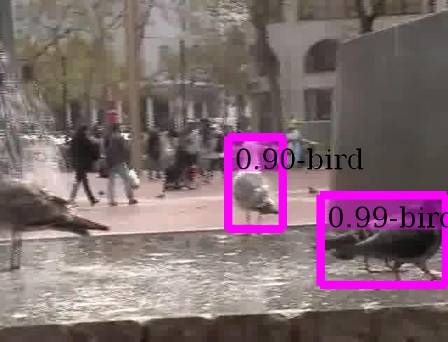}&\includegraphics[trim=0 40 0 70,clip,width=0.24\linewidth]{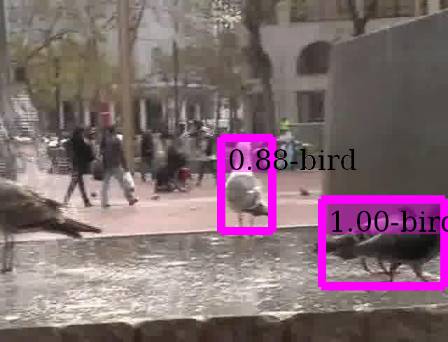} \\
    \textbf{MemNet}&\includegraphics[trim=0 40 0 70,clip,width=0.24\linewidth]{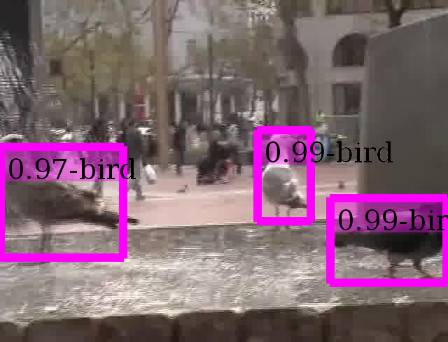}&\includegraphics[trim=0 40 0 70,clip,width=0.24\linewidth]{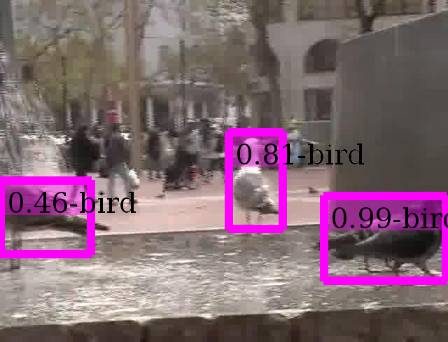}&\includegraphics[trim=0 40 0 70,clip,width=0.24\linewidth]{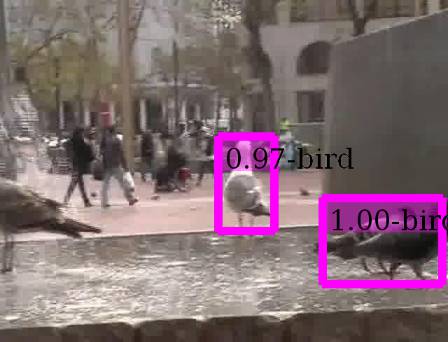} \\
    \textbf{ClockNet}&\includegraphics[trim=0 40 0 70,clip,width=0.24\linewidth]{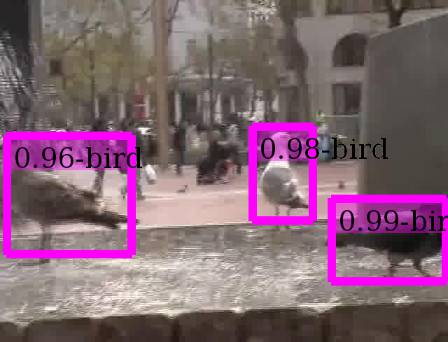}&\includegraphics[trim=0 40 0 70,clip,width=0.24\linewidth]{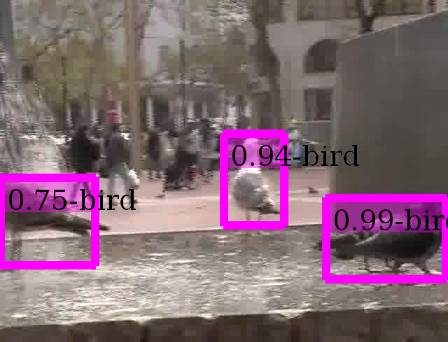}&\includegraphics[trim=0 40 0 70,clip,width=0.24\linewidth]{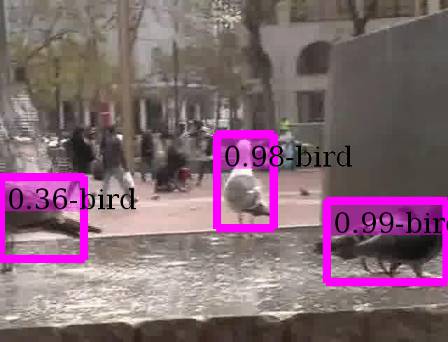} \\

			\textbf{RFCN}&\includegraphics[width=0.24\linewidth]{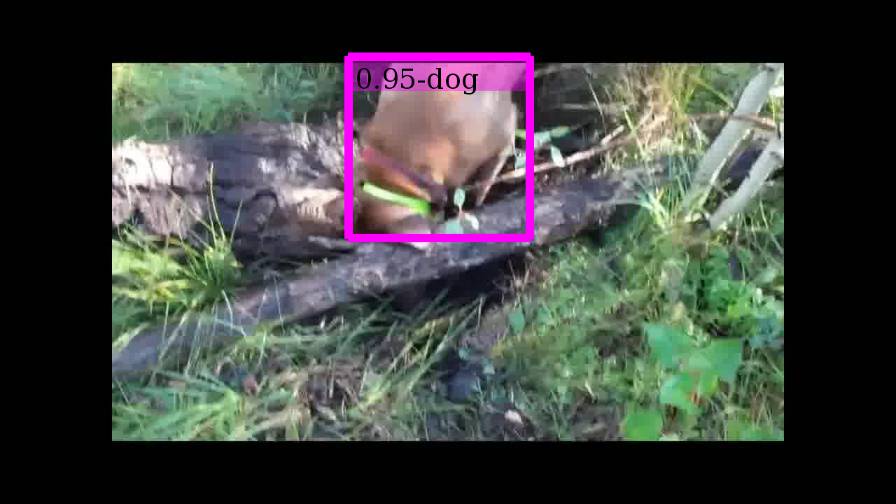}& \includegraphics[width=0.24\linewidth]{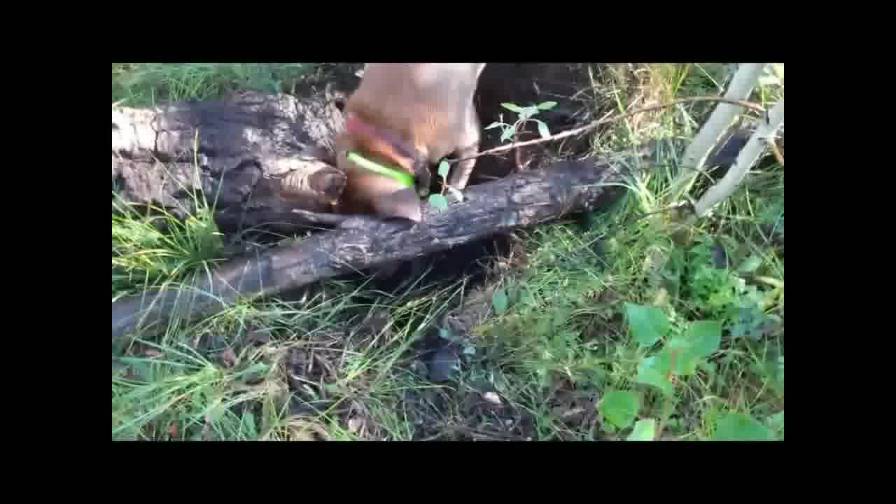}& \includegraphics[width=0.24\linewidth]{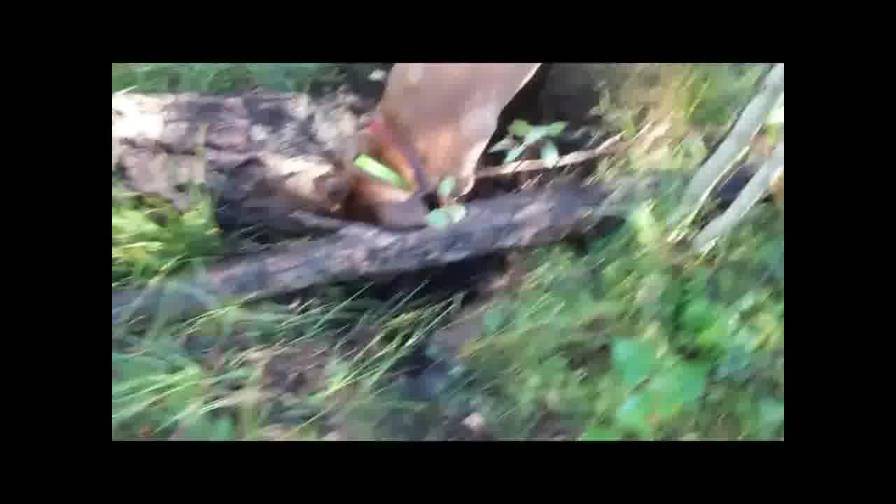} \\
			\textbf{MemNet}&\includegraphics[width=0.24\linewidth]{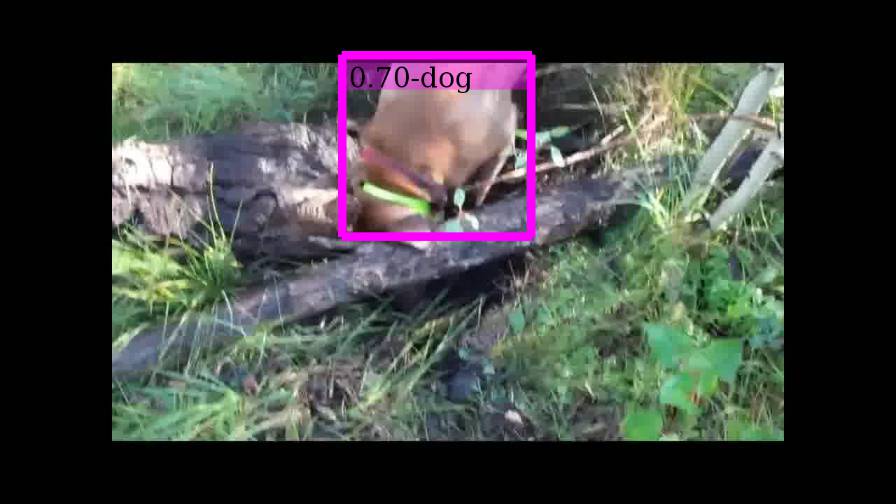}& \includegraphics[width=0.24\linewidth]{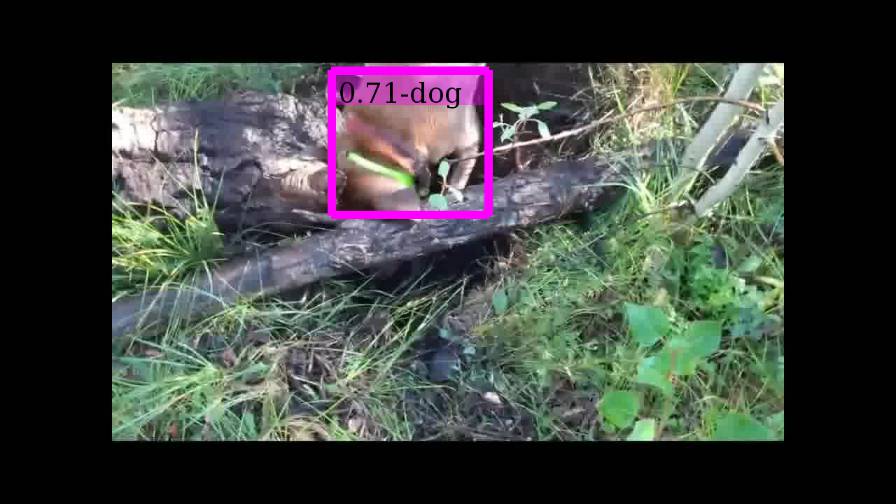}& \includegraphics[width=0.24\linewidth]{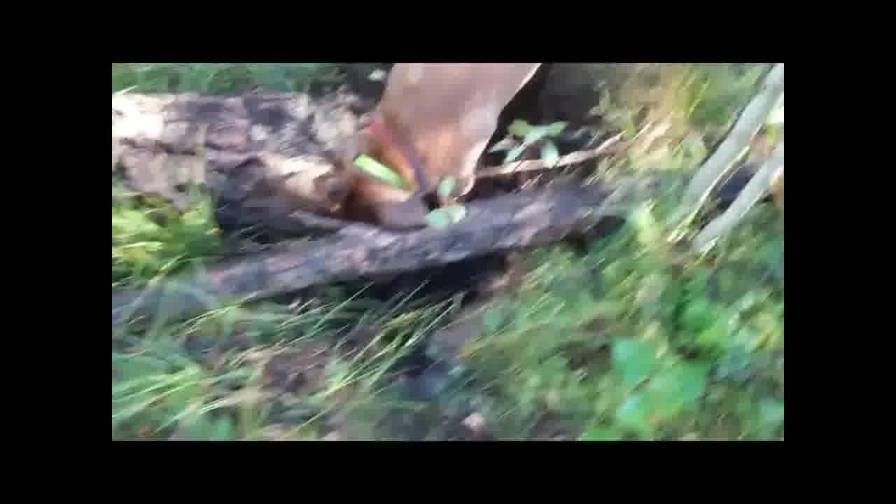} \\
			\textbf{ClockNet}&\includegraphics[width=0.24\linewidth]{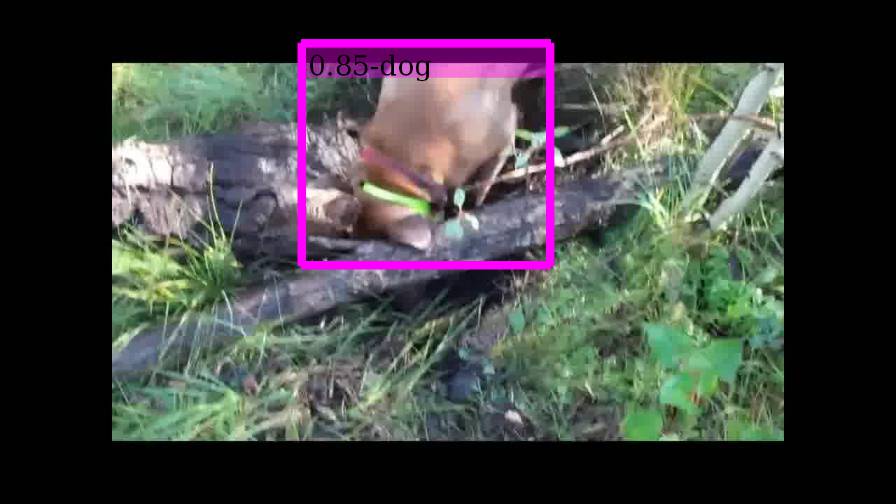}& \includegraphics[width=0.24\linewidth]{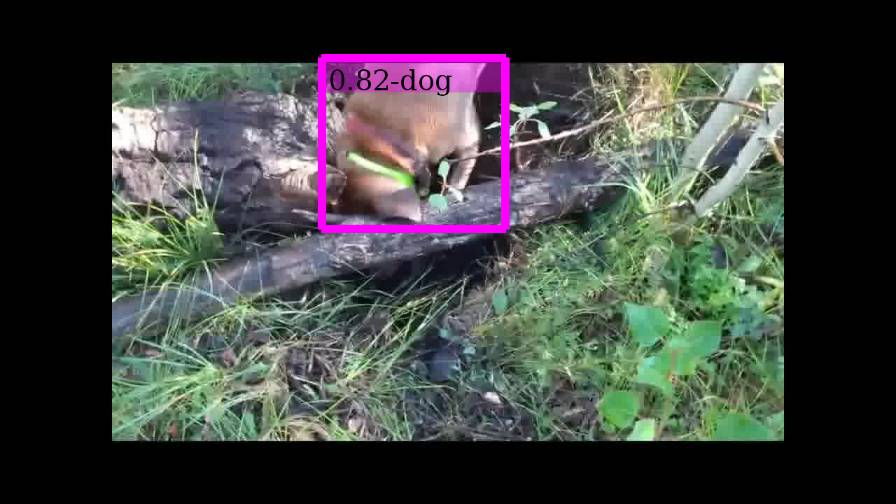} & \includegraphics[width=0.24\linewidth]{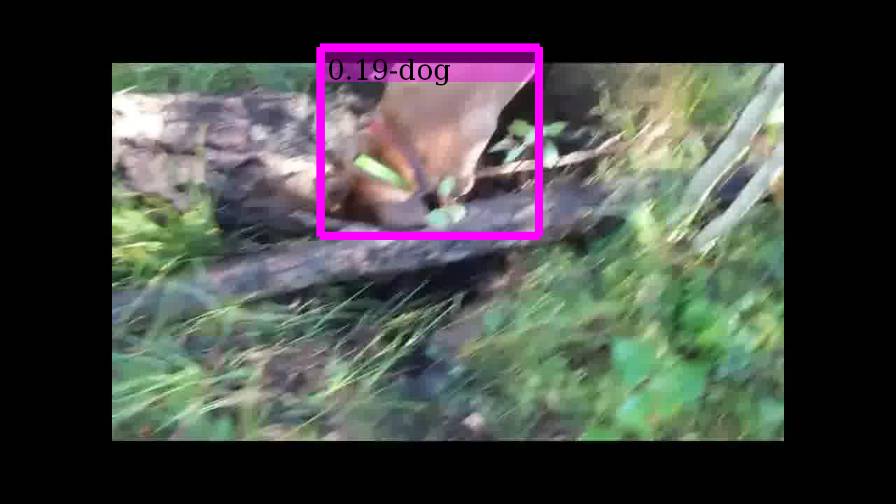} \\

			\textbf{RFCN}&\includegraphics[width=0.24\linewidth]{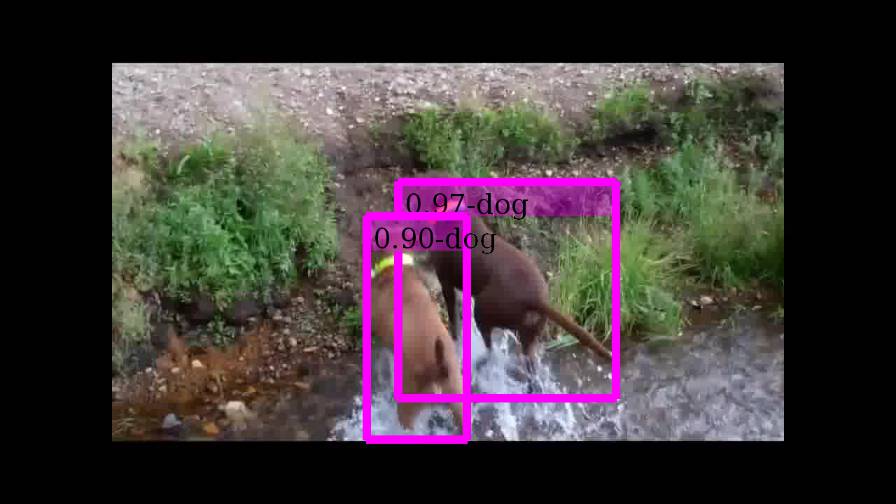}& \includegraphics[width=0.24\linewidth]{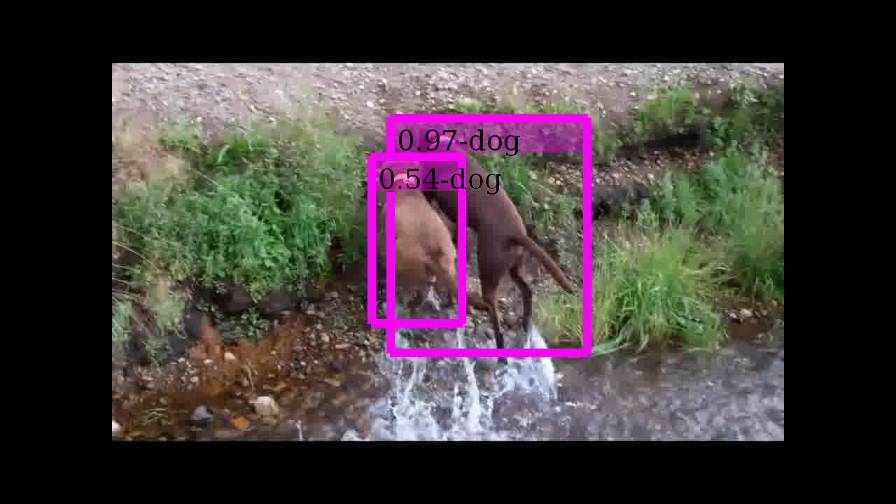}& \includegraphics[width=0.24\linewidth]{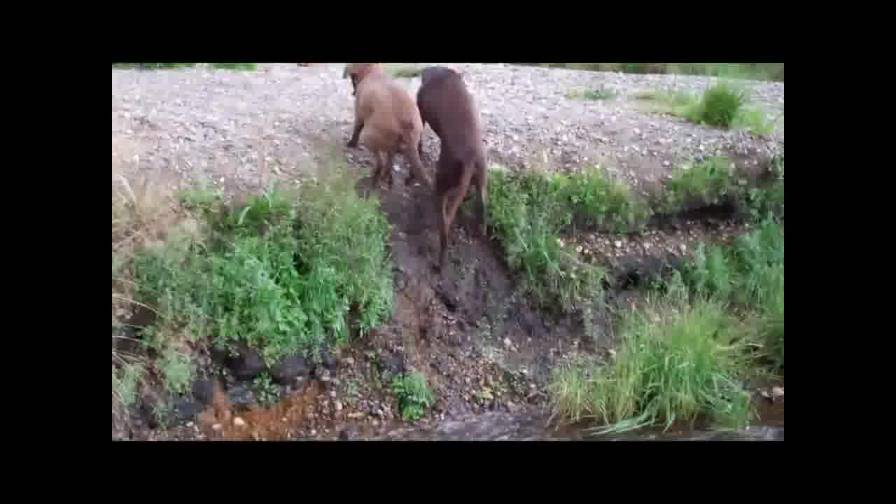} \\
			\textbf{MemNet}&\includegraphics[width=0.24\linewidth]{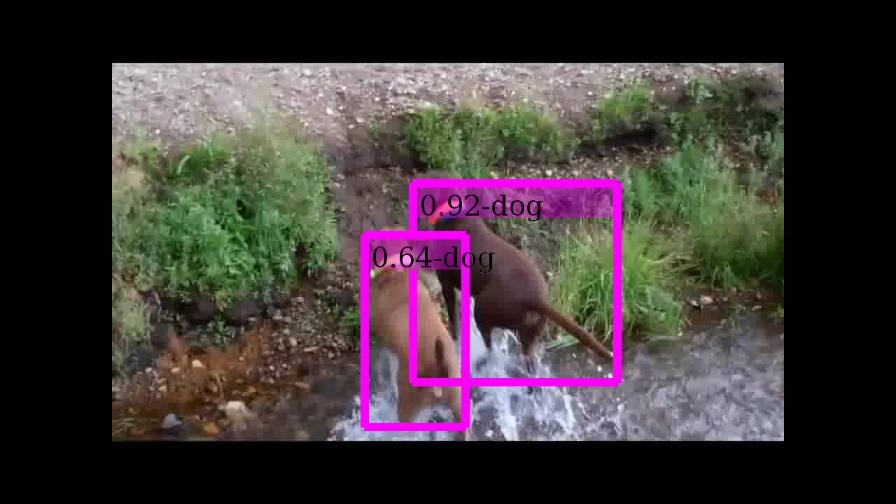}& \includegraphics[width=0.24\linewidth]{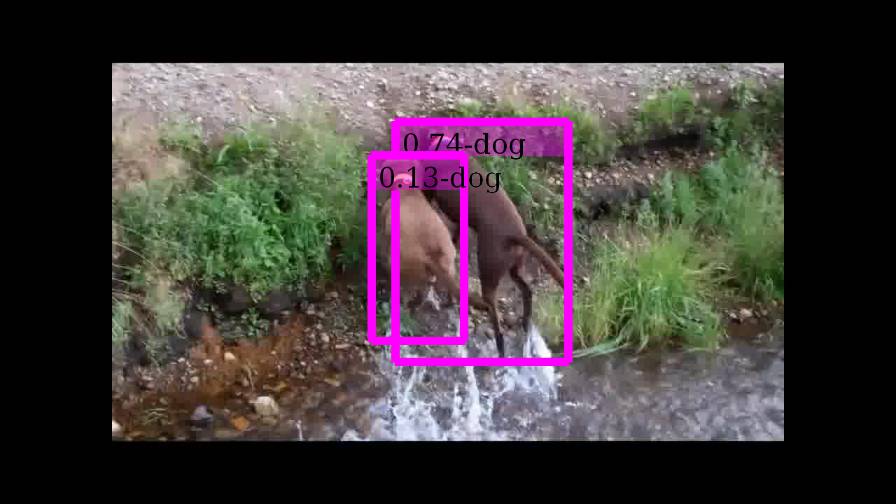}& \includegraphics[width=0.24\linewidth]{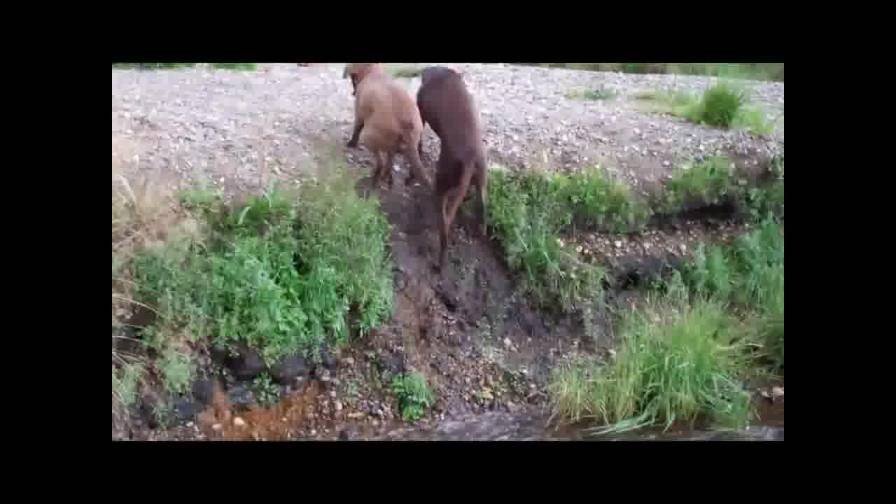} \\
			\textbf{ClockNet}&\includegraphics[width=0.24\linewidth]{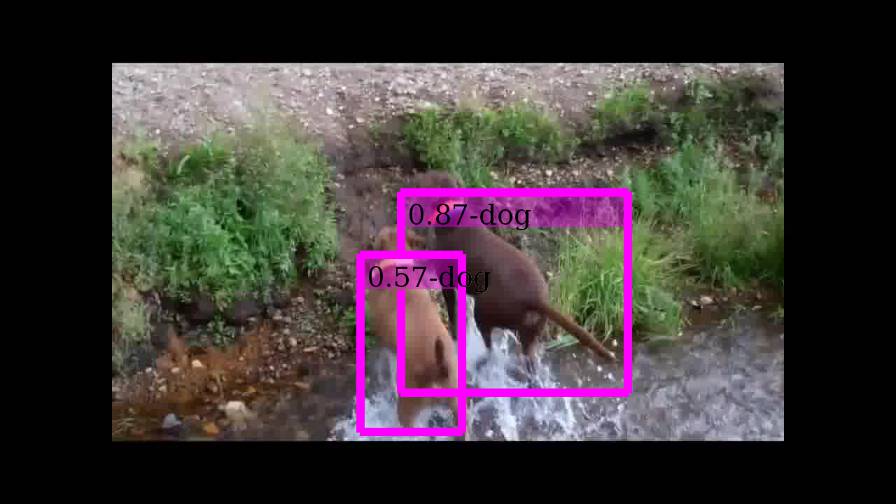}& \includegraphics[width=0.24\linewidth]{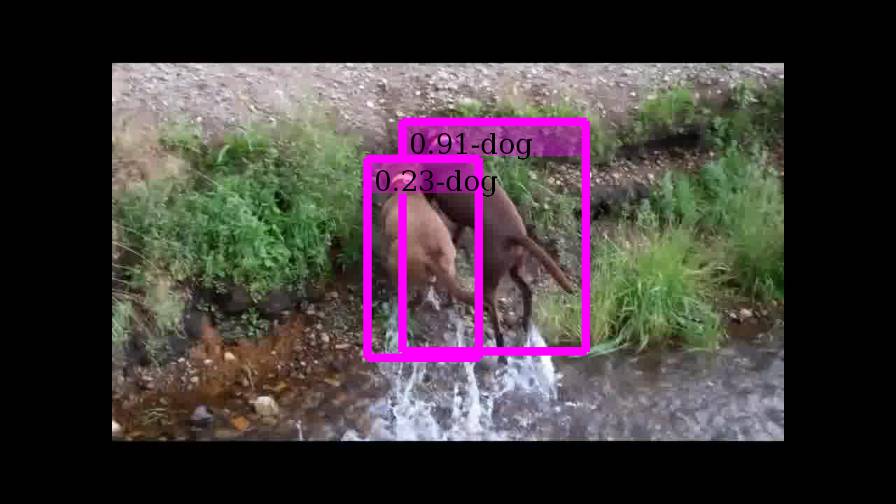} & \includegraphics[width=0.24\linewidth]{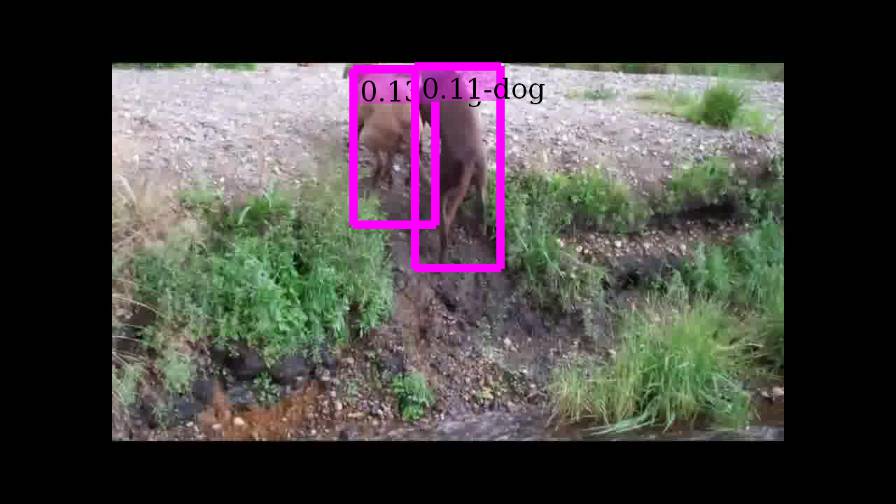} \\
  \end{tabular}
	\caption{Qualitative results of MemNet, ClockNet and the RFCN baseline}
	\label{fig:qual_res_imagenetVID_supmat}
\end{figure}

Figure~\ref{fig:qual_res_imagenetVID_supmat} illustrates additional qualitative results of our memory networks and the RFCN baseline~\cite{thvu:dai2016r}, as in Figure 7(a-b) of the main paper.

\section{Feature anticipation}
\label{sec:anticipation}
\begin{figure}[h!]\centering
  \includegraphics[trim=100 900 100 100, clip, width=1.0\linewidth]{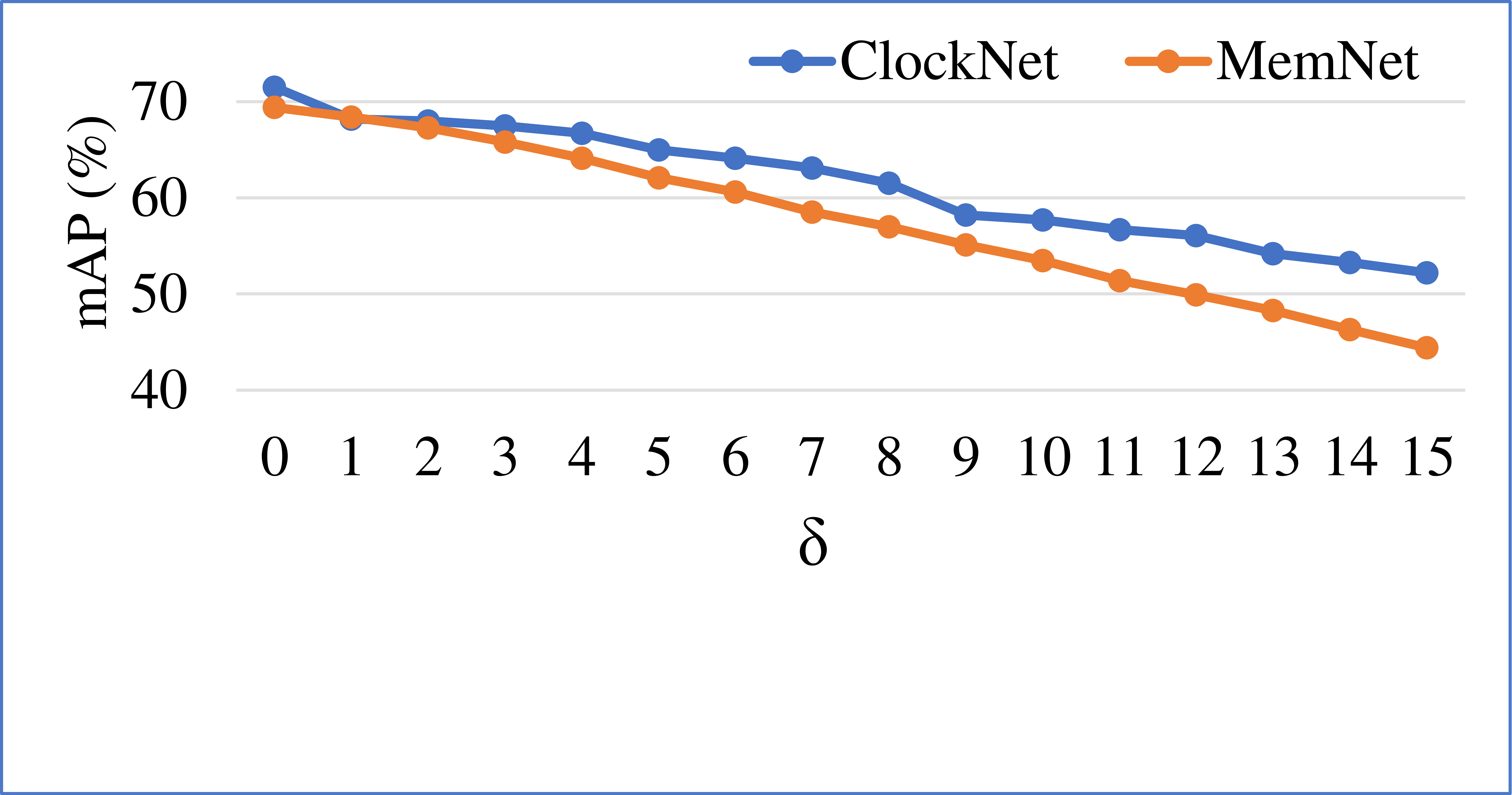}
  \vspace{-1.0cm}
	\caption{Mean AP of MemNet and ClockNet with respect to different anticipation lengths $\delta$.}
	\label{fig:exp_feature_anticipation}
\end{figure}

In this section, we complement the feature anticipation experiment conducted in Section 5.3 of the main paper.  Due to space limitations, we only showed a plot for feature propagation (Figure 7d) but not for anticipation.  Here we provide the same figure for feature anticipation, see Figure~\ref{fig:exp_feature_anticipation}.  We can observe that ClockNet is more efficient than MemNet in anticipating features over long distances, which is similar to our observations for feature propagation.

\bibliographystyle{splncs}
\bibliography{myshortstrings,mem_rfcn_biblist}

\begin{thebibliography}{10}

\bibitem{Cutting_1986}
Cutting, J.E.:
\newblock Perception with an Eye for Motion.
\newblock MIT Press (1986)

\bibitem{Gibson_1979}
Gibson, J.:
\newblock The Ecological Approach to Visual Perception.
\newblock Houghton Mifflin (1979)

\bibitem{Ellis_1938}
Ellis, W.:
\newblock A Source Book of {G}estalt Psychology.
\newblock Routledge (1938)

\bibitem{Wertheimer_1938}
Wertheimer, M.:
\newblock Laws of organization in perceptual forms.
\newblock Psycologische Forschung \textbf{3} (1938)

\bibitem{sam:Russakovsky15a}
Russakovsky, O., Deng, J., Su, H., Krause, J., Satheesh, S., Ma, S., Huang, Z.,
  Karpathy, A., Khosla, A., Bernstein, M., Berg, A.C., Fei-Fei, L.:
\newblock {ImageNet Large Scale Visual Recognition Challenge}.
\newblock IJCV \textbf{115}(3) (2015)  211--252

\bibitem{sam:abuelhaija16a}
Abu-El-Haija, S., Kothari, N., Lee, J., Natsev, P., Toderici, G., Varadarajan,
  B., Vijayanarasimhan, S.:
\newblock {YouTube-8M: A Large-Scale Video Classification Benchmark}.
\newblock arXiv:1609.08675 (2016)

\bibitem{sam:Lin17b}
Lin, T.Y., Doll\'{a}r, P., Girshick, R., He, K., Hariharan, B., Belongie, S.:
\newblock {Feature Pyramid Networks for Object Detection}.
\newblock In: CVPR. (2017)

\bibitem{sam:Lin17a}
Lin, T.Y., Goyal, P., Girshick, R., He, K., Doll\'{a}r, P.:
\newblock {Focal Loss for Dense Object Detection}.
\newblock In: ICCV. (2017)

\bibitem{sam:Liu16a}
Liu, W., Anguelov, D., Erhan, D., Szegedy, C., Reed, S., Fu, C.Y., Berg, A.C.:
\newblock {SSD: Single Shot MultiBox Detector}.
\newblock In: ECCV. (2016)

\bibitem{sam:Ren15a}
Ren, S., He, K., Girshick, R., Sun, J.:
\newblock {Faster R-CNN: Towards Real-Time Object Detection with Region
  Proposal Networks}.
\newblock In: NIPS. (2015)

\bibitem{sam:Kang16a}
Kang, K., Ouyang, W., Li, H., Wang, X.:
\newblock {Object Detection from Video Tubelets with Convolutional Neural
  Networks}.
\newblock In: CVPR. (2016)

\bibitem{sam:Feichtenhofer17c}
Feichtenhofer, C., Pinz, A., Zisserman, A.:
\newblock {Detect to Track and Track to Detect}.
\newblock In: ICCV. (2017)

\bibitem{Han_etal_2016}
Han, W., Khorrami, P., Paine, T.L., Ramachandran, P., Babaeizadeh, M., Shi, H.,
  Li, J., Yan, S., Huang, T.S.:
\newblock Seq-{N}{M}{S} for video object detection.
\newblock CoRR \textbf{1602.08465} (2016)

\bibitem{sam:Zhu17b}
Zhu, X., Xiong, Y., Dai, J., Yuan, L., Wei, Y.:
\newblock {Flow-Guided Feature Aggregation for Video Object Detection}.
\newblock ICCV (2017)

\bibitem{thvu:dai2016r}
Dai, J., Li, Y., He, K., Sun, J.:
\newblock {R-FCN: Object detection via region-based fully convolutional
  networks}.
\newblock In: NIPS. (2016)

\bibitem{sam:Koutnik14a}
Koutn\'{i}k, J., Greff, K., Gomez, F., Schmidhuber, J.:
\newblock {A Clockwork RNN}.
\newblock In: ICML. (2014)

\bibitem{sam:Shelhamer16a}
Shelhamer, E., Rakelly, K., Hoffman, J., Darrell, T.:
\newblock {Clockwork Convnets for Video Semantic Segmentation}.
\newblock In: Video Semantic Segmentation Workshop at ECCV. (2016)

\bibitem{Wang_etal_2017}
Wang, X., Shrivastava, A., Gupta, A.:
\newblock {A}-{F}ast-{R}{C}{N}{N}: {H}ard positive generation via adversary for
  object detection.
\newblock In: CVPR. (2017)  3039--3048

\bibitem{Shrivastava_etal_2016}
Shrivastava, A., Gupta, A., Girshick, R.:
\newblock Training region-based object detectors with online hard example
  mining.
\newblock In: CVPR. (2016)  761--769

\bibitem{sam:Donahue15a}
Donahue, J., Hendricks, L.A., Guadarrama, S., Rohrbach, M., Venugopalan, S.,
  Saenko, K., Darrell, T.:
\newblock {Long-term Recurrent Convolutional Networks for Visual Recognition
  and Description}.
\newblock In: CVPR. (2015)

\bibitem{sam:Srivastava15a}
Srivastava, N., Mansimov, E., Salakhutdinov, R.:
\newblock {Unsupervised Learning of Video Representations using LSTMs}.
\newblock In: ICML. (2015)

\bibitem{sam:Hochreiter97a}
Hochreiter, S., Schmidhuber, J.:
\newblock {Long Short-Term Memory}.
\newblock Neural Computation \textbf{9} (1997)  1735--1780

\bibitem{sam:Tran15a}
Tran, D., Bourdev, L., Fergus, R., Torresani, L., Paluri, M.:
\newblock {Learning Spatiotemporal Features with 3D Convolutional Networks}.
\newblock In: ICCV. (2015)

\bibitem{sam:Yao15a}
Yao, L., Torabi, A., Cho, K., Ballas, N., Pal, C., Larochelle, H., Courville,
  A.:
\newblock {Describing Videos by Exploiting Temporal Structure}.
\newblock In: ICCV. (2015)

\bibitem{Wang_Gupta_2015}
Wang, X., Gupta, A.:
\newblock Unsupervised learning of visual representations using videos.
\newblock In: ICCV. (2015)  2794--2802

\bibitem{sam:Simonyan14a}
Simonyan, K., Zisserman, A.:
\newblock {Two-Stream Convolutional Networks for Action Recognition in Videos}.
\newblock In: NIPS. (2014)

\bibitem{sam:Feichtenhofer16a}
Feichtenhofer, C., Pinz, A., Zisserman, A.:
\newblock {Convolutional Two-Stream Network Fusion for Video Action
  Recognition}.
\newblock In: CVPR. (2016)

\bibitem{sam:Feichtenhofer17a}
Feichtenhofer, C., Pinz, A., Wildes, R.P.:
\newblock {Spatiotemporal Multiplier Networks for Video Action Recognition}.
\newblock In: CVPR. (2017)

\bibitem{sam:Zhu17a}
Zhu, X., Xiong, Y., Dai, J., Yuan, L., Wei, Y.:
\newblock {Deep Feature Flow for Video Recognition}.
\newblock In: CVPR. (2017)

\bibitem{thvu:dosovitskiy2015flownet}
Dosovitskiy, A., Fischer, P., Ilg, E., Hausser, P., Hazirbas, C., Golkov, V.,
  van~der Smagt, P., Cremers, D., Brox, T.:
\newblock {FlowNet: Learning Optical Flow With Convolutional Networks}.
\newblock In: ICCV. (2015)

\bibitem{sam:Han16a}
Han, W., Khorrami, P., Paine, T.L., Ramachandran, P., Babaeizadeh, M., Shi, H.,
  Li, J., Yan, S., Huang, T.S.:
\newblock {Seq-NMS for Video Object Detection}.
\newblock CoRR \textbf{abs/1602.08465} (2016)

\bibitem{sam:Kundu16a}
Kundu, A., Vineet, V., Koltun, V.:
\newblock {Feature Space Optimization for Semantic Video Segmentation}.
\newblock In: CVPR. (2016)

\bibitem{Luc_etal_2017}
Luc, P., Neverova, N., Couprie, C., Verbeek, J., LeCun, Y.:
\newblock {Predicting Deeper into the Future of Semantic Segmentation}.
\newblock In: ICCV. (2017)

\bibitem{Vondrick_etal_2016}
Vondrick, C., Pirsiavash, H., Torralba, A.:
\newblock {Anticipating Visual Representations From Unlabeled Video}.
\newblock In: CVPR. (2016)

\bibitem{sam:Yang16a}
Yang, F., Choi, W., Lin, Y.:
\newblock {Exploit All the Layers: Fast and Accurate CNN Object Detector With
  Scale Dependent Pooling and Cascaded Rejection Classifiers}.
\newblock In: CVPR. (2016)

\bibitem{sam:Mathieu16a}
Mathieu, M., Couprie, C., Lecun, Y.:
\newblock {Deep Multi Scale Video Prediction Beyond Mean Square Error}.
\newblock (2016)

\bibitem{sam:Jaderberg15a}
Jaderberg, M., Simonyan, K., Zisserman, A., Kavukcuoglu, K.:
\newblock {Spatial Transformer Networks}.
\newblock In: NIPS. (2015)

\bibitem{sam:Pinheiro14a}
Pinheiro, P.O., Collobert, R.:
\newblock {Recurrent Convolutional Neural Networks for Scene Labeling} (2014)

\bibitem{sam:Liang15a}
Liang, M., Hu, X.:
\newblock {Recurrent Convolutional Neural Network for Object Recognition}.
\newblock In: CVPR. (2015)

\bibitem{sam:Cho14a}
Cho, K., van Merrienboer, B., Gulcehre, C., Bahdanau, D., Bougares, F.,
  Schwenk, H., Bengio, Y.:
\newblock {Learning Phrase Representations using RNN Encoder-Decoder for
  Statistical Machine Translation}.
\newblock In: EMNLP. (2014)

\bibitem{sam:Redmon16a}
Redmon, J., Divvala, S., Girshick, R., Farhadi, A.:
\newblock {You Only Look Once: Unified, Real-Time Object Detection}.
\newblock In: CVPR. (2016)

\bibitem{sam:Redmon17a}
Redmon, J., Farhadi, A.:
\newblock {YOLO9000: Better, Faster, Stronger}.
\newblock In: CVPR. (2017)

\bibitem{thvu:he2016deep}
He, K., Zhang, X., Ren, S., Sun, J.:
\newblock {Deep Residual Learning for Image Recognition}.
\newblock In: CVPR. (2016)

\bibitem{thvu:chen2016deeplab}
Chen, L.C., Papandreou, G., Kokkinos, I., Murphy, K., Yuille, A.L.:
\newblock Deeplab: Semantic image segmentation with deep convolutional nets,
  atrous convolution, and fully connected crfs.
\newblock arXiv preprint arXiv:1606.00915 (2016)

\bibitem{thvu:nair2010rectified}
Nair, V., Hinton, G.E.:
\newblock Rectified linear units improve restricted boltzmann machines.
\newblock In: ICML. (2010)

\bibitem{thvu:MxNet}
Chen, T., Li, M., Li, Y., Lin, M., Wang, N., Wang, M., Xiao, T., Xu, B., Zhang,
  C., Zhang, Z.:
\newblock Mxnet: {A} flexible and efficient machine learning library for
  heterogeneous distributed systems.
\newblock CoRR (2015)

\bibitem{thvu:kar2016adascan}
Kar, A., Rai, N., Sikka, K., Sharma, G.:
\newblock Adascan: Adaptive scan pooling in deep convolutional neural networks
  for human action recognition in videos.
\newblock arXiv preprint arXiv:1611.08240 (2016)

\end{thebibliography}

\end{document}